\pdfoutput=1

\documentclass[11pt]{article}

\usepackage[]{EMNLP2022}

\usepackage{times}
\usepackage{latexsym}
\usepackage{amsmath}

\usepackage[T1]{fontenc}

\usepackage[utf8]{inputenc}

\usepackage{microtype}

\usepackage{inconsolata}

\usepackage{graphicx}
\usepackage{booktabs}
\usepackage{multirow}
\usepackage{multicol}
\usepackage{adjustbox}
\usepackage{subfigure}
\usepackage{hyperref}
\usepackage{amssymb}
\usepackage{xspace,mfirstuc,tabulary}
\usepackage{hhline}
\usepackage[ruled,noend]{algorithm2e}
\usepackage{amsmath, bm}
\usepackage{color}
\usepackage{subfigure}

\usepackage{cleveref}
\crefname{section}{§}{§§}
\Crefname{section}{§}{§§}
\crefname{figure}{Figure}{Figure}
\Crefname{figure}{Figure}{Figure}
\crefname{table}{Table}{Table}
\Crefname{table}{Table}{Table}
\newcommand\ourdataset{COPEN\xspace}

\newcommand\BERTsmall{BERT$_{\textsc{SMALL}}$\xspace}
\newcommand\BERTmedium{BERT$_{\textsc{MEDIUM}}$\xspace}
\newcommand\BERTbase{BERT$_{\textsc{BASE}}$\xspace}
\newcommand\BERTlarge{BERT$_{\textsc{LARGE}}$\xspace}
\newcommand\Rbase{RoBERTa$_{\textsc{BASE}}$\xspace}
\newcommand\GPTbase{GPT-2$_{\textsc{BASE}}$\xspace}
\newcommand\GPTmedium{GPT-2$_{\textsc{MEDIUM}}$\xspace}
\newcommand\GPTlarge{GPT-2$_{\textsc{LARGE}}$\xspace}
\newcommand\GPTxl{GPT-2$_{\textsc{XL}}$\xspace}
\newcommand\GPTNeobase{GPT-Neo$_{\textsc{125M}}$\xspace}
\newcommand\BARTbase{BART$_{\textsc{BASE}}$\xspace}
\newcommand\Tsmall{T5$_{\textsc{SMALL}}$\xspace}
\newcommand\Tbase{T5$_{\textsc{BASE}}$\xspace}
\newcommand\Tlarge{T5$_{\textsc{LARGE}}$\xspace}
\newcommand\Txl{T5$_{\textsc{3B}}$\xspace}
\newcommand\Txxl{T5$_{\textsc{11B}}$\xspace}

\title{\ourdataset: Probing Conceptual Knowledge in Pre-trained Language Models}

\author{Hao Peng$^{1,2}$\thanks{\quad Equal contribution}\hspace{0.5em}, Xiaozhi Wang$^{1,2*}$, 
Shengding Hu$^{1,2}$, Hailong Jin$^{1,2}$, Lei Hou$^{1,2}$\thanks{\quad Corresponding author: L.Hou}\hspace{0.5em}, \\
\textbf{Juanzi Li$^{1,2}$, Zhiyuan Liu$^{1,2}$, Qun Liu$^{3}$} \\ 
$^1$Department of Computer Science and Technology, BNRist; \\
$^2$KIRC, Institute for Artificial Intelligence,\\
Tsinghua University, Beijing, 100084, China\\
$^3$Huawei Noah’s Ark Lab \\
{\tt \{peng-h21, wangxz20\}@mails.tsinghua.edu.cn}
}

\begin{document}
\maketitle

\begin{abstract}
    Conceptual knowledge is fundamental to human cognition and knowledge bases. However, existing knowledge probing works only focus on evaluating factual knowledge of pre-trained language models (PLMs) and ignore conceptual knowledge. Since conceptual knowledge often appears as implicit commonsense behind texts, designing probes for conceptual knowledge is hard. Inspired by knowledge representation schemata, we comprehensively evaluate conceptual knowledge of PLMs by designing three tasks to probe whether PLMs organize entities by conceptual similarities, learn conceptual properties, and conceptualize entities in contexts, respectively. For the tasks, we collect and annotate $24$k data instances covering $393$ concepts, which is \ourdataset, a COnceptual knowledge Probing bENchmark. Extensive experiments on different sizes and types of PLMs show that existing PLMs systematically lack conceptual knowledge and suffer from various spurious correlations. We believe this is a critical bottleneck for realizing human-like cognition in PLMs. \ourdataset and our codes are publicly released at \url{https://github.com/THU-KEG/COPEN}.

\end{abstract}

\section{Introduction}

Pre-trained language models (PLMs) have achieved superior performance on most NLP tasks requiring substantial world knowledge~\citep{qiu2020pre,han2021pre}. It is interesting and meaningful to \textit{probe} the extent and scope of world knowledge within PLMs. Existing knowledge probing works have evaluated PLMs' knowledge about entities~\citep{broscheit-2019-investigating,tenney-etal-2019-bert} and their relations~\cite{DBLP:conf/emnlp/PetroniRRLBWM19, DBLP:journals/tacl/JiangXAN20, DBLP:conf/emnlp/RobertsRS20}, i.e., factual knowledge, but ignore conceptual knowledge.

\begin{figure}
    \centering
    \includegraphics[width=0.9\linewidth]{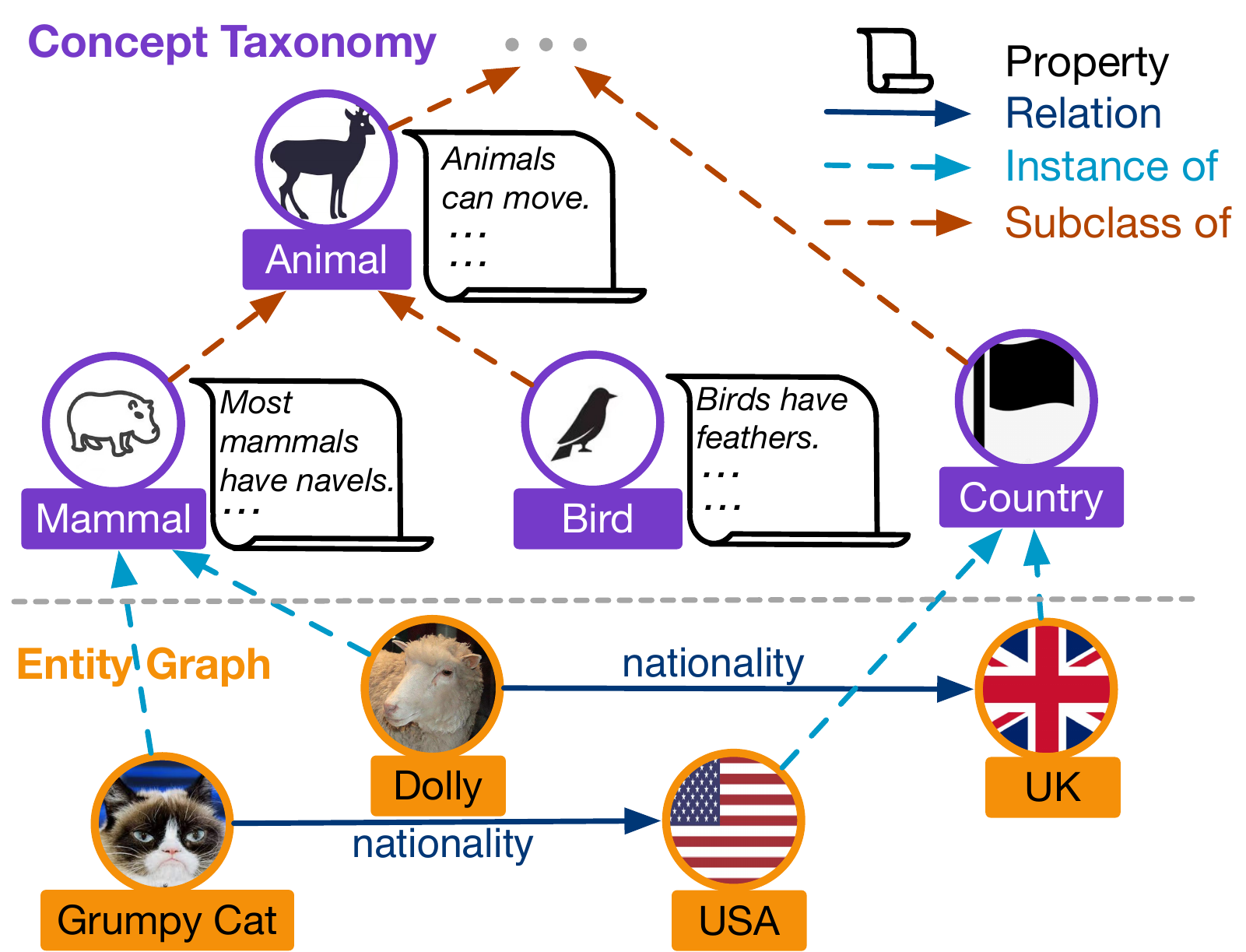}
    \caption{An example knowledge graph. Entities are organized by concepts through the \texttt{Instance of} relation and concepts are organized into a taxonomy through the \texttt{Subclass of} relation. Each concept has certain properties. Existing work only probes factual knowledge in entity graphs, ignoring conceptual knowledge in the concept taxonomy and \texttt{Instance of} relation.}
    \label{fig:example}
\end{figure}

Conceptual knowledge, especially the abstraction ability, is fundamental to all kinds of human cognition~\citep{carey1991knowledge,Collins2014KnowledgeIP} including language processing~\citep{Waxman1995WordsAI,10.3389/fpsyg.2014.00506}. Just as the quote of psychologist Gregory Murphy, \textit{concepts are the glue that holds our mental world together}~\citep{murphy2004big}. Moreover, knowledge bases~\citep{yago,auer2007dbpedia,vrandevcic2012wikidata} organize massive entities via concept taxonomies as illustrated in \cref{fig:example}, which enable broad applications~\citep{lv-etal-2018-differentiating,zhou-etal-2021-kacc}. Therefore, probing whether PLMs have human-like conceptual knowledge is necessary in knowledge probing.

Inspired by the conceptual schema in knowledge representations~\citep{sowa1976conceptual,DBLP:journals/internet/DeckerMHFKBEH00,mcguinness2004owl,DBLP:books/daglib/0036180}, we comprehensively evaluate the conceptual knowledge of PLMs by asking three questions:  Do PLMs organize entities by conceptual similarities? Do PLMs know the properties of concepts? Can PLMs correctly conceptualize entities in contexts? In this paper, we design three probing tasks for these questions: (1) The \textbf{conceptual similarity judgment (CSJ)} task studies whether PLMs organize entities by conceptual similarities, which is the basis of understanding concepts.
Given a query entity, CSJ requires PLMs to choose the most conceptually similar entity among candidate entities. For example, in \cref{fig:example}, given \texttt{Dolly} as the query entity, although \texttt{UK} has a direct relation and more co-occurrences with it, PLMs should choose \texttt{Grumpy Cat}.
(2) The \textbf{conceptual property judgment (CPJ)} task probes whether PLMs have the knowledge of conceptual properties, which are the generic abstractions of factual knowledge.
Given a statement about a specific property, such as ``\textit{have feathers}'', CPJ requires PLMs to judge whether it is true for a specific concept and also a concept chain, which evaluates whether PLMs understand the property transitivity among a chain of hierarchical concepts.
(3) The \textbf{conceptualization in contexts (CiC)} task evaluates the abilities of PLMs to correctly conceptualize entities within contexts. Given an entity mentioned in a specific context, PLMs are required to choose the most appropriate concept in a concept taxonomy according to its context. CiC requires not only disambiguating entity mentions, but also distinguishing superordinate and subordinate concepts. For instance, given the context ``\textit{Dolly is running on the grassland}'', PLMs should conceptualize \texttt{Dolly} as an \texttt{Animal} since there is no enough evidence for \texttt{Mammal}.

Based on the above considerations, we construct a conceptual knowledge probing benchmark, \ourdataset, which contains a concept taxonomy with $446$ concepts and high-quality data of $24$K instances for the three probing tasks. The concept taxonomy is curated by experts based on DBpedia~\citep{auer2007dbpedia} and Wikidata~\citep{vrandevcic2014wikidata} to form a well-defined hierarchy and cover broad entities. The data instances for three tasks are collected by aligning entities in Wikidata and sentences in GenericsKB~\citep{DBLP:journals/corr/abs-2005-00660}, Wikipedia\footnote{\url{https://en.wikipedia.org/}}, and Simple Wikipedia\footnote{\url{https://simple.wikipedia.org/}} into the concept taxonomy and then manually annotated by crowd-sourcing annotators.

We conduct extensive experiments on \ourdataset to evaluate various widely-used language models (LMs), which include three types: masked LMs~\citep{DBLP:conf/naacl/DevlinCLT19,DBLP:journals/corr/abs-1907-11692}, autoregressive LMs~\citep{radford2019language,gpt-neo}, and sequence-to-sequence LMs~\citep{DBLP:conf/acl/LewisLGGMLSZ20,DBLP:journals/jmlr/RaffelSRLNMZLL20}. We conduct the experiments in three settings: (1) zero-shot probing, which reformulates the probing tasks into pre-training objectives and lets PLMs score answers without any training~\citep{DBLP:conf/emnlp/PetroniRRLBWM19}; (2) linear probing, which only tunes additional linear classification heads and uses them to handle probing tasks with the frozen representations produced by PLMs; (3) fine-tuning, which tunes all the PLM parameters. Experiments show that existing PLMs achieve non-trivial performance but still significantly underperform ordinary persons on all three probing tasks. Further analyses show that PLMs suffer from spurious correlations like word co-occurrences and out-of-context predictions, and increasing model scale brings marginal improvements. 

To summarize, our contributions are three-fold: (1) We propose to probe PLMs for conceptual knowledge, which has long been ignored, and design three probing tasks inspired by the knowledge representation works. (2) We construct \ourdataset, a probing benchmark containing high-quality concept taxonomy and probes. (3) We empirically show that existing PLMs systematically lack conceptual knowledge and analyze the reasons. We hope our benchmark and findings could facilitate further research on concept-aware PLMs and human-like language understandings.

\section{\ourdataset Benchmark}

\begin{figure*}[t!]
    \centering
    \includegraphics[width=0.98\linewidth]{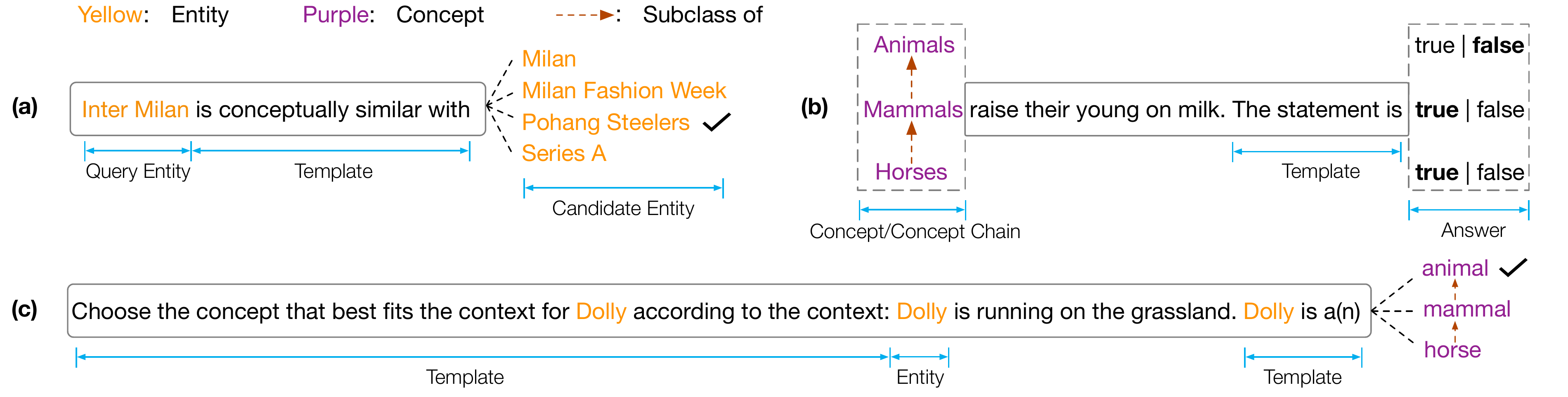}
    \caption{Examples for casting the data of three probing tasks into natural language prompts in zero-shot probing. The names of entities or concepts are the text looked up in Wikidata using their IDs. In Figure (b), \textbf{texts in bold} (true or false) denote answers. In Figure (b) and (c), the concept chain is \texttt{Horse} --> \texttt{Mammal} --> \texttt{Animal}. In Figure (c), for entities with multiple concept chains, each concept will be scored independently by PLMs, i.e., the PLMs make concept-level predictions only. There is no dedicated chain selection procedure. }
    \label{fig:method}
\end{figure*}

In this session, we introduce our \ourdataset benchmark, including the construction of the concept taxonomy (\cref{sec:concept_taxonomy}) and the datasets for three probing tasks (\cref{sec:CSJ,sec:CPJ,sec:CiC}). More construction and annotation details are shown in \cref{sec:appendix_copen}.

\subsection{COPEN Concept Taxonomy}
\label{sec:concept_taxonomy}

Designing the three probing tasks takes inspiration from concept schemata in knowledge representations~\citep{DBLP:journals/internet/DeckerMHFKBEH00,mcguinness2004owl}, which are widely used in knowledge graphs~\citep{yago,auer2007dbpedia,vrandevcic2012wikidata}. In general, it uses the \texttt{instance of} relation to link the entities (specific instances) into abstract concepts, and uses the \texttt{subclass of} relation to organize the concepts into a taxonomy. Each concept has certain properties describing it as the example shown in \cref{fig:example}.

To support probing dataset construction, we manually curate a concept taxonomy based on DBpedia~\citep{auer2007dbpedia} and Wikidata~\citep{vrandevcic2014wikidata} in $3$ steps: (1) Obtain a basic taxonomy from DBpedia. We extract the frequent concepts of DBpedia, which are the concepts with at least $5$ instances, and keep the \texttt{subclass of} relations between them. (2) Align DBpedia and Wikidata. For each DBpedia concept, we manually find its equivalent Wikidata item and then use the \texttt{subclass of} (\texttt{P279}) relations in Wikidata to expand the concept taxonomy and use the \texttt{instance of} (\texttt{P31}) relations to link massive Wikidata entities into the concepts. (3) Simplify the taxonomy. We further remove some unusual concepts to simplify the taxonomy by the guidance from Schema.org~\citep{guha2016schema}. For example, \texttt{Person} is a sub-concept of \texttt{Animal}, \texttt{Eukaryote}, and \texttt{Species} in DBpedia, which is reasonable but inconvenient for real-world applications. Following Schema.org, we set \texttt{Person} as a top-level concept in the taxonomy. Finally, we achieve a tree-structure concise concept taxonomy, which contains $446$ concepts covering $45$ million Wikidata entities. There are $23$ top-level concepts, and we use $11$ of them and their sub-concepts for constructing training and development datasets as well as the other concepts for the testing datasets.


\subsection{Conceptual Similarity Judgment}
\label{sec:CSJ}

The conceptual similarity judgment (CSJ) task is a multiple-choice classification task, which probes whether PLMs organize entities by conceptual similarities, i.e., whether PLMs learn the \texttt{instance of} relation. Given a query entity, CSJ requires PLMs to choose the most conceptually similar entity (\texttt{instance of} the same superordinate concept) among some candidates. As in \cref{fig:method} (a), PLMs should choose \texttt{Pohang Steelers} for \texttt{Inter Milan} since they are both football clubs, although \texttt{Milan} and \texttt{Inter Milan} co-occur more frequently. The conceptual similarity here is similar to the \textit{cohyponym} relation in lexical semantics~\citep{cruse1986lexical}, which has been shown to be distinct from but easily influenced by spurious co-occurrence associations~\citep{simlex-999}. Thus we need to control the influence of co-occurrences to get faithful results.

\begin{table}[t!]
    \centering
    \small
    \begin{tabular}{llrrr}
\toprule
     & & \multicolumn{1}{c}{Train} & \multicolumn{1}{c}{Dev} & \multicolumn{1}{c}{Test} \\ \midrule
     \multirow{2}{*}{CSJ} & \#Instance & $4{,}462$ & $1{,}116$ & $3{,}909$   \\
     & \#Concept  & $84$ & $84$ & $90$ \\ 
     \midrule
    \multirow{2}{*}{CPJ} & \#Instance & $3{,}274$  & $823$  & $4{,}758$ \\
     & \#Concept  & $215$   & $195$ & $178$ \\ 
     \midrule
    \multirow{2}{*}{CiC} & \#Instance & $2{,}888$  & $722$  & $2{,}368$ \\
     & \#Concept  & $193$  & $184$  & $155$ \\ 
     \bottomrule
    \end{tabular}

    \caption{\ourdataset data statistics for three probing tasks.}
    \label{tab:statistics}
\end{table}

\paragraph{Data Collection}
The data for CSJ is collected in two steps: (1) Automatic collection. We first sample $174$ concepts that are not subordinates to each other. Then we retrieve $50$ Wikidata entities most frequently showing up in the Wikipedia corpus for each concept, and then build data instances by combining them. Each instance consists of a query entity, an answer entity of the same concept, and $20$ distractor entities, among which $5$ are hard distractors of concepts sharing superordinates with the concept of query entity. To check the data quality, we sample $200$ instances and find little noise. (2) Co-occurrence-based filtering. To reduce the influence of co-occurrences, we need to filter out the instances that can be easily solved with co-occurrences. \citet{word-embedding-compare} show that Glove word embedding~\citep{glove} contains rich word co-occurrence information but limited cohyponym knowledge. Hence we use it to filter out instances with higher word similarity between the query and answer entity than distractor entities. 
We finally get $9{,}487$ instances, each including a query entity and $21$ candidate entities. The statistics of data subsets are shown in \cref{tab:statistics}.

\subsection{Conceptual Property Judgment}
\label{sec:CPJ}
The conceptual property judgment (CPJ) task is a binary sentence classification task, which probes whether PLMs know the \textit{properties} of concepts. Given a statement describing a certain conceptual property, PLMs are required to judge whether it is true. For example in \cref{fig:method} (b), PLMs should predict ``true'' for the statement instance \textit{Mammals raise their young on milk}. 

Besides evaluating CPJ at instance level, which reflects the PLMs' knowledge about properties for different individual concepts, we also set a \textbf{chain-level} evaluation, in which a PLM correctly judges a property if and only if it correctly judges the property for every concept in a \textit{concept chain}. As the example in \cref{fig:method}~(b), a concept chain is a chain of concepts connected with the \texttt{subclass of} relation in order. The chain-level evaluation evaluates whether PLMs understand the transitivity of conceptual properties. It means that a property holds for a concept also holds for its subordinate concepts, but may not hold for its superordinate concepts like the case in \cref{fig:method} (b).

\paragraph{Data Collection}
The data for CPJ is collected in two steps: (1) Automatic collection. For each concept in our taxonomy, we align it with the statements of GenericsKB~\citep{DBLP:journals/corr/abs-2005-00660}, a high-quality knowledge base for naturally occurring generic statements, by lexical matching so as to get positive instances. Then we replace the concept mention with other concept names to obtain negative instances. (2) Human annotation. To ensure data quality, we invite annotators to check whether the instances are correctly labeled, grammatically correct, and describing concept properties. All annotators are well-trained and pass a qualification before annotation. We finally get $8{,}855$ instances for CPJ and the statistics of data subsets are shown in Table~\ref{tab:statistics}. Additionally,  
the final test data includes $102$ concept chains and corresponding properties used for chain-level evaluation.

\subsection{Conceptualization in Contexts}
\label{sec:CiC}

The conceptualization in contexts (CiC) task is a multiple-choice classification task, which probes whether PLMs can correctly conceptualize entities within contexts. Given an entity mentioned in a specific sentence, PLMs are required to choose the most appropriate concept among a concept chain, which is a chain of concepts connected with the \texttt{subclass of} relation in order. This requires PLMs to understand the \texttt{subclass of} relation and capture the subtle differences of different-level concepts in a hierarchy. For example in \cref{fig:method} (c), given the sentence \textit{Dolly is running on the grassland.} and a concept chain \texttt{Horse} --> \texttt{Mammal} --> \texttt{Animal}, PLMs shall choose \texttt{Animal} for \texttt{Dolly} since the context do not support more fine-grained concepts. Sometimes the entity is of multiple concept chains, for example, \texttt{Jimmy Carter} is both a \texttt{Writer} and a \texttt{Politician}, which additionally requires PLMs to disambiguate.

\paragraph{Data Collection}
The data for CiC is collected in two steps: (1) Sentence collection. For each concept, we first retrieve $10$ Wikidata entities most frequently showing up in the Wikipedia corpus. Among the retrieved entities, we only keep the entities linked with the concept chains containing more than one concepts and collect $5$ sentences for each of them from Wikipedia and SimpleWiki, which provides various contexts for conceptualization. A sentence, together with an entity mentioned in the sentence and concept chains of the entity, constitutes an instance. (2) Human annotation. We then organize crowd-sourcing annotation to obtain the labels. All annotators are well-trained and qualified. We finally get $5{,}978$ instances for CiC and the statistics of data subsets are shown in Table~\ref{tab:statistics}.

\begin{table*}[t!]
    \centering
    \small
    \resizebox{\linewidth}{!}{%
    \begin{tabular}{l|rrr|rrr|rrr|rrr}
    \toprule
    \multirow{3}{*}{Model} &
    \multicolumn{3}{c|}{CSJ} & 
    \multicolumn{6}{c|}{CPJ} &
    \multicolumn{3}{c}{CiC} \\
    \cmidrule{5-10} 
    & & & & \multicolumn{3}{c|}{Instance-Level} & \multicolumn{3}{c|}{Chain-Level} & & & \\ 
    \cmidrule{2-13} 
    & ZP & LP & FT & ZP & LP & FT & ZP & LP & FT & ZP & LP & FT  \\
    \midrule
    Random &  $4.8$ & $4.8$ & $4.8$ & 
    $50.0$ & $50.0$ & $50.0$ & 
    $7.2$ & $7.2$ & $7.2$ & 
    $27.7$ & $27.7$ & $27.7$ \\ 
    \midrule
\BERTbase&$\mathbf{20.3}$&$\mathbf{16.1}_{\textsc{0.21}}$&$27.3_{\textsc{0.86}}$&$49.4$&$61.6_{\textsc{0.28}}$&$68.1_{\textsc{0.98}}$&$\mathbf{22.5}$&$\mathbf{24.2}_{\textsc{1.22}}$&$\mathbf{23.2}_{\textsc{1.22}}$&$37.6$&$34.3_{\textsc{0.59}}$&$49.5_{\textsc{0.60}}$\\
\Rbase&$15.5$&$12.0_{\textsc{0.21}}$&$22.3_{\textsc{0.51}}$&$49.2$&$61.9_{\textsc{0.13}}$&$72.0_{\textsc{0.54}}$&$21.6$&$13.1_{\textsc{1.67}}$&$18.3_{\textsc{1.22}}$&$31.4$&$30.0_{\textsc{1.98}}$&$52.6_{\textsc{1.02}}$\\
\GPTbase&$7.9$&$4.3_{\textsc{0.24}}$&$20.1_{\textsc{0.23}}$&$51.5$&$64.8_{\textsc{1.14}}$&$70.4_{\textsc{0.72}}$&$14.7$&$14.4_{\textsc{0.92}}$&$20.3_{\textsc{2.01}}$&$32.3$&$34.5_{\textsc{2.08}}$&$\mathbf{54.2}_{\textsc{0.12}}$\\
\GPTNeobase&$7.9$&$11.0_{\textsc{0.20}}$&$18.3_{\textsc{0.42}}$&$52.2$&$62.2_{\textsc{0.21}}$&$68.2_{\textsc{0.62}}$&$22.5$&$15.0_{\textsc{2.01}}$&$19.0_{\textsc{2.81}}$&$32.6$&$39.6_{\textsc{0.93}}$&$47.4_{\textsc{0.25}}$\\
\BARTbase&$14.4$&$8.4_{\textsc{0.10}}$&$21.0_{\textsc{0.50}}$&$48.7$&$58.5_{\textsc{0.27}}$&$68.2_{\textsc{0.86}}$&$20.6$&$10.5_{\textsc{1.22}}$&$16.7_{\textsc{0.80}}$&$33.6$&$\mathbf{43.7}_{\textsc{1.19}}$&$51.3_{\textsc{1.56}}$\\
\Tbase&$15.2$&$4.9_{\textsc{0.21}}$&$\mathbf{27.9}_{\textsc{0.60}}$&$\mathbf{55.9}$&$\mathbf{66.9}_{\textsc{0.25}}$&$\mathbf{72.5}_{\textsc{0.28}}$&$22.5$&$18.0_{\textsc{0.46}}$&$18.0_{\textsc{3.95}}$&$\mathbf{42.3}$&$24.7_{\textsc{0.66}}$&$53.2_{\textsc{0.18}}$\\
    \midrule
    Human &$79.5$ & $79.5$ & $79.5$ & $91.4$ & $91.4$ & $91.4$ & $91.2$ & $91.2$ & $91.2$ & $85.6$ & $85.6$ & $85.6$ \\ 
    \bottomrule
    \end{tabular}
    }
    \caption{Accuracies (\%) of various PLMs on the three tasks using different probing methods. ZP: Zero-shot probing. LP: Linear probing. FT: Fine-tuning. 
    LP and FT results are $\mathrm{Mean}_{\mathrm{standard~deviation}}$ over three random trials. Human performance is obtained by ordinary people trained with a few instances.}
    \label{tab:main_experiment}
\end{table*}

\section{Evaluation Setup}
We introduce the various widely-used PLMs investigated in our experiments (\cref{sec:plms}) and the three adopted probing methods (\cref{sec:probing_method}).

\subsection{Investigated PLMs}
\label{sec:plms}
We investigate three mainstream types of PLMs:
(1) \textbf{Masked LM}, including BERT~\citep{DBLP:conf/naacl/DevlinCLT19}, which is pre-trained with the bidirectional masked language modeling and next sentence prediction objectives, and RoBERTa~\citep{DBLP:journals/corr/abs-1907-11692}, which is a robustly optimized version of BERT.
(2) \textbf{Autoregressive LM}, including GPT-2~\citep{radford2019language}, which is pre-trained with the unidirectional left-to-right language modeling objective, and GPT-Neo~\citep{gpt-neo}, which adopts the same objective but improves some implementation details.
(3) \textbf{Sequence-to-sequence LM}, which adopts the encoder-decoder architecture. This type includes BART~\citep{DBLP:conf/acl/LewisLGGMLSZ20}, which is pre-trained with the text infilling and sentence permutation objectives, and T5~\citep{DBLP:journals/jmlr/RaffelSRLNMZLL20}, which is pre-trained with the span-corruption objective and multiple downstream tasks.

In \cref{sec:experiment}, we report the results of the frequently-used \textsc{BASE} versions of these PLMs, and results for the other versions are shown in appendix~\ref{sec:appendix_exp}.

\subsection{Probing Method}
\label{sec:probing_method}

\textbf{Zero-Shot Probing} reformulates probing tasks to the format of pre-training language modeling objectives~\citep{DBLP:journals/corr/abs-2107-13586} so that PLMs can do these tasks without any training. It is widely adopted by knowledge probing work~\citep{DBLP:conf/emnlp/PetroniRRLBWM19,tenney-etal-2019-bert} since it prevents PLMs from learning new knowledge from training data so that the achieved performance reflects PLMs' intrinsic knowledge. Hence the performance of zero-shot probing is commonly interpreted as the \textit{lower bound} of PLMs' knowledge~\citep{DBLP:journals/tacl/JiangXAN20}.

As illustrated in \cref{fig:method}, for each data instance of the three probing tasks, we cast its choices into natural language prompts by filling them into manually designed templates, and then let PLMs score the prompts by the likelihood of language modeling. The choice with the highest score is regarded as the predicted answer of PLMs. Some implementation details like taking which parts of the prompts into scoring calculation may influence the PLMs' performance. We search these details with preliminary trials and only report the performance of the best configuration in experiments.

\textbf{Linear Probing} adds an additional shallow linear classifier on top of the output contextualized representations of PLMs, and only trains the additional classifier while keeping the PLMs' parameters fixed. Since the model capacity of the shallow linear classifier is too limited to fit the tasks, the achieved performance shall mainly come from the knowledge in the PLMs' representations~\citep{DBLP:conf/iclr/AlainB17}. Hence linear probing is widely used in knowledge probing~\citep{tenney2019you,hewitt-manning-2019-structural}. 

\textbf{Fine-Tuning} is the standard method to adapt PLMs to downstream tasks, which trains all the PLMs' parameters on the training data with task-specific objectives. Considering the strong model capacity of the PLMs, PLMs will inevitably fit the probing tasks through the information in training data rather than only resort to their intrinsic knowledge. Hence  the fine-tuning performance shall serve as an \textit{upper bound} of the PLMs' conceptual knowledge in our experiments.

For CSJ and CiC, we take the filled prompts of identical templates in zero-shot probing as inputs and train PLMs with the cross-entropy loss. For CPJ, we take the property statements as inputs and use the binary cross entropy loss.

More detailed implementations about three probing methods are shown in \cref{sec:appendix_experiment}.

\section{Experiment and Analysis}
\label{sec:experiment}
We first introduce the overall results in \cref{sec:exp_overall} and conduct detailed analyses on the three probing tasks (\cref{sec:CSJ_exp,sec:CPJ_exp,sec:CiC_exp}), respectively. We then analyze the performance at different model scales (\cref{sec:analyze_scale}). 
More observations and discussions on experimental results are placed in \cref{sec:appendix_discuss}. 

\subsection{Overall Results}
\label{sec:exp_overall}

The overall experimental results are shown in \cref{tab:main_experiment}, from which we can observe that:
(1) All the PLMs can achieve non-trivial (better than random guess) performance on all the probing tasks with zero-shot probing or linear probing, which indicates that existing PLMs capture a certain conceptual knowledge with pre-training on massive texts. (2) However, even with fine-tuning, all PLMs' accuracies are still well below human performance, which urges further efforts on concept-aware pre-training. 
(3) The accuracies of PLMs using different types of pre-training objectives are generally on the same level. It suggests that any existing pre-training objective has no special advantages in understanding concepts and further improvements may come from targeted pre-training design. We provide some analyses in the following sections to help targeted concept-aware PLMs development.

\begin{table}
    \centering
    \small
        \begin{tabular}{lrr}
    \toprule
    Model &  Hard Distractor & Easy Distractor \\
    \midrule
    \BERTbase & $25.1$ & $15.7$\\
    \Rbase &  $25.3$ & $15.7$ \\
    \GPTbase &  $21.1$ & $17.0$\\
    \GPTNeobase & $20.7$ & $17.1$\\ 
    \BARTbase & $24.2$ & $16.0$ \\
    \Tbase & $24.6$ & $15.9$ \\
    \bottomrule
    \end{tabular} %
    \caption{Mean reciprocal ranks (\%) for hard distractors and easy distractors on CSJ in zero-shot probing results of various PLMs. Larger values for higher ranks.}
    \label{tab:task1_mrr}
\end{table}

\subsection{Conceptual Similarity Judgment}
\label{sec:CSJ_exp}
We analyze the predictions and performance of various PLMs on CSJ, and find that:

\paragraph{PLMs better distinguish coarse-grained concepts.}
As mentioned in \cref{sec:CSJ}, among $20$ distractor entities, $5$ of them are hard distractors of concepts sharing superordinates with the concept of the query entity, and the others are easy distractors. For example, if the query entity is of \texttt{Mammal} concept, the entities of \texttt{Bird} concept are hard distractors and the entities of \texttt{Country} concept are easy distractors. \cref{tab:task1_mrr} shows the mean reciprocal ranks of these two kinds of distractors. We can see that the hard distractors are significantly ranked higher than easy distractors, which indicates that PLMs generally better distinguish coarse-grained concepts, such as telling the differences between \texttt{Animal} and \texttt{Country}, but fail in distinguishing fine-grained concepts. It suggests that future methods should focus more on how to capture the subtle differences between fine-grained concepts.

\subsection{Conceptual Property Judgment}
\label{sec:CPJ_exp}
We analyze the error cases on CPJ and find that:

\paragraph{Conceptual transitivity challenges PLMs.} 
\cref{tab:main_experiment} shows that PLMs can achieve high instance-level accuracies, but all perform poorly in the chain-level evaluation. It suggests that PLMs can relatively well recall the properties for individual concepts like recalling the facts about entities in factual knowledge probing, but hardly understand the hierarchical relations of concepts and the property transitivity. It suggests that further PLM works should not only focus on better memorizing knowledge but also consider how to better organize knowledge.

\paragraph{PLMs have \textit{conceptual hallucination}.}
It has been observed that PLMs frequently generate nonsensical and unfaithful outputs, which are factually incorrect, and previous work~\citep{DBLP:conf/emnlp/RohrbachHBDS18, hallucination-in-neural-nlg, DBLP:journals/corr/abs-2202-03629} dubs this phenomenon as \textit{hallucination}. In our experiments, we observe that many PLMs' failure cases on CPJ task can be described as \textit{conceptual hallucination}, i.e., PLMs hallucinate that concepts have certain properties while they actually do not. As shown in Table~\ref{tab:falsepos}, the errors of most PLMs are generally mainly from making false positive predictions, i.e., taking false conceptual property statements as true. It suggests that PLMs tend to hallucinate the false conceptual properties as true rather than cannot recall the true conceptual properties, which is interesting and we further explore whether there are certain spurious correlations causing this.

\begin{table}[t!]
    \centering
    \small
    \resizebox{\columnwidth}{!}{%
    \begin{tabular}{rrrrrr}
    \toprule
    BERT & RoBERTa & GPT-2 & GPT-Neo & BART & T5  \\ 
    \midrule 
    $78.0$ & $72.5$ & $64.6$ & $52.5$ & $65.9$ & $58.3$ \\
    \bottomrule
    \end{tabular}%
    }
    \caption{Percentage (\%) of false positive predictions among all incorrect predictions in fine-tuning results of various PLMs on the CPJ dataset.}
    \label{tab:falsepos}
\end{table}

\begin{figure}
    \centering
    \includegraphics[width=0.8\linewidth]{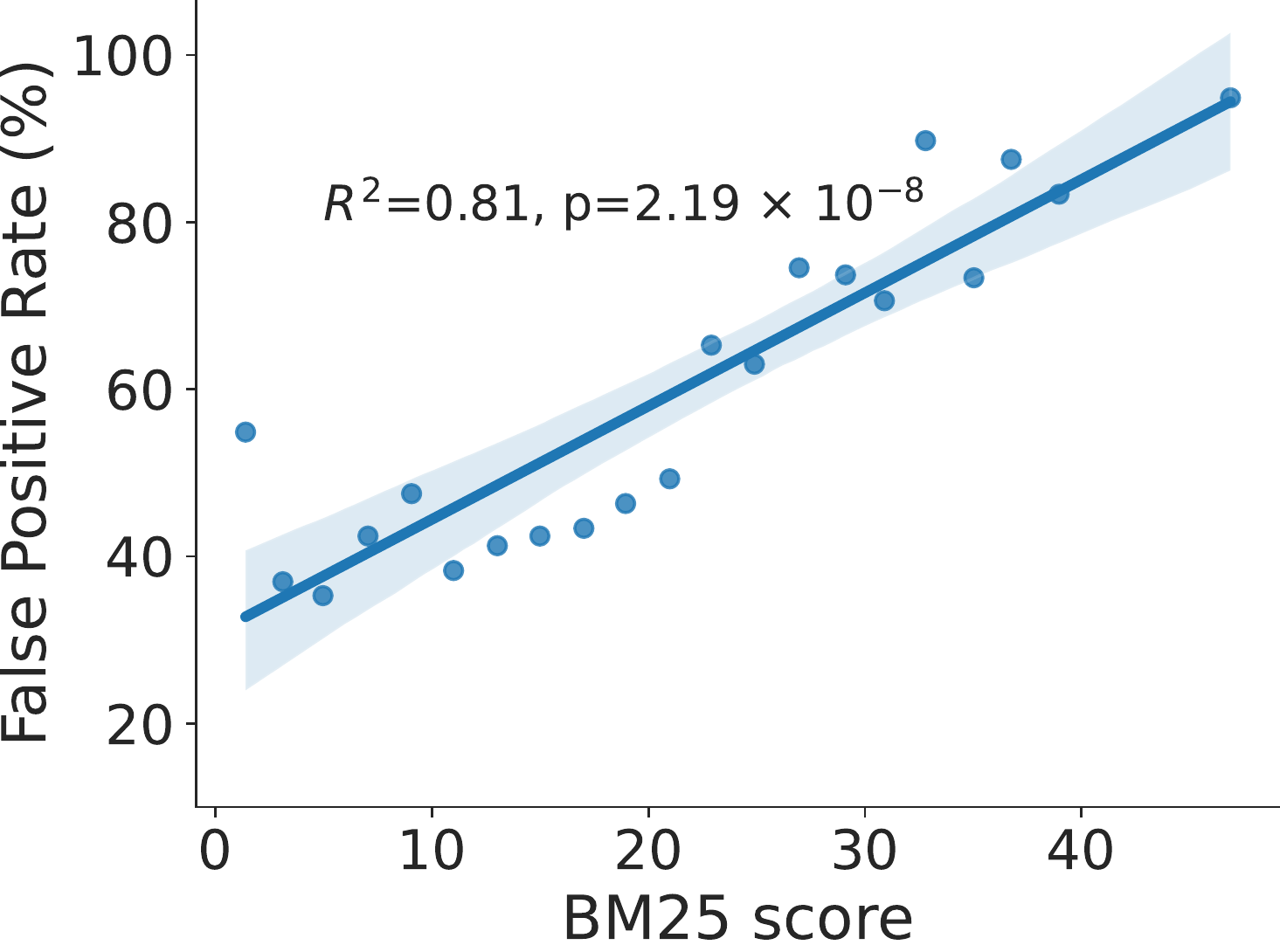}
    \caption{The false positive rate of BERT's fine-tuning results on CPJ negative instances with different BM25 scores. Results of other PLMs are left in \cref{sec:appendix_hallucination}.}
    \label{fig:error_rate}
\end{figure}

\paragraph{Word co-occurrence causes conceptual hallucination.}
We hypothesize that the word co-occurrence in the pre-training corpora causes PLMs' conceptual hallucination. For example, if a PLM has seen the text ``\textit{The temple's Jufu Hall was included in the 1998 World \textbf{Monuments} Watch by the World \textbf{Monuments} Fund (WMF) ...preservation of the painted \textbf{decoration}}''\footnote{\url{https://en.wikipedia.org/wiki/Temple\_of\_Agriculture}}, it may be more likely to predict the statement ``\textit{Monuments are used for decoration}'' as true. We empirically find pieces of evidence supporting this hypothesis. For each CPJ instance, to assess the word co-occurrence in pre-training corpora, we retrieve the most similar document of it from Wikipedia, which is a widely-used corpus in pre-training, with the BM25~\citep{robertson1995okapi} algorithm implemented in Whoosh~\citep{mchaput_2016}, and use the BM25 score of the top one of retrieved documents as the indicator of this CPJ instance's word co-occurrence rate in pre-training corpus. We divide the negative instances of CPJ dataset into different subsets by their BM25 scores and observe the false positive rate of BERT's fine-tuning predictions on them. The results are plotted in \cref{fig:error_rate}, from which we can see that the false positive prediction rates, indicating conceptual hallucination, have strong positive correlations to the BM25 scores, indicating word co-occurrence. This suggests that the conceptual hallucination of PLMs comes from capturing the spurious correlations of word co-occurrence in pre-training, and further pre-training work shall explore to fix it.

\subsection{Conceptualization in Contexts}
\label{sec:CiC_exp}
We analyze the error cases on CiC and find that:

\paragraph{PLMs conceptualize entities over-relying on memories.}
In CiC, we find that if we remove the contexts, PLMs can still predict a possibly correct concept, which is similar to previous works~\citep{DBLP:conf/emnlp/PetroniRRLBWM19, DBLP:conf/emnlp/RobertsRS20,DBLP:conf/acl/CaoLHSYLXX20} showing that PLMs memorize a certain knowledge about entities' types. We dub these predictions \textit{out-of-context predictions}, which can be regarded as the PLMs' memories obtained in pre-training. What we evaluate in CiC is the in-context conceptualization abilities rather than the memorized knowledge about the concepts of entities, which is evaluated by CSJ. Hence relying on the memories and making out-of-context predictions are wrong for handling CiC. However, as shown in \cref{tab:out_of_context}, in most of the error cases, PLMs wrongly conceptualize the entities within contexts as the default out-of-context predictions. It demonstrates that PLMs conceptualize entities by over-relying on memories rather than understanding the contexts, which reflects the lack of genuine conceptualization abilities. We encourage future works to study whether the memories inhibit learning to conceptualize during pre-training.

\begin{table}[t!]
    \centering
    \small
    \resizebox{\columnwidth}{!}{%
    \begin{tabular}{rrrrrr}
    \toprule
    BERT & RoBERTa & GPT-2 & GPT-Neo & BART & T5  \\ 
    \midrule 
    $72.9$ & $75.9$ & $76.7$ & $60.4$ & $71.8$ & $59.2$ \\
    \bottomrule
    \end{tabular}%
    }
    \caption{Percentage (\%) of out-of-context predictions among all incorrect predictions in zero-shot probing results of various PLMs on the CiC dataset.
    }
    \label{tab:out_of_context}
\end{table}

\begin{table*}
    \centering
    \small
    \begin{tabular}{lll}
    \toprule
    \textbf{Error Type} & \textbf{Context} & \textbf{Concept Chains} \\ 
    \midrule
    Disambiguation & He was nominated by President  & \texttt{Person} --> \underline{\texttt{BusinessPerson}} \\ 
    $29.0\%$ &  \textit{Jimmy Carter} to the court.
    & \texttt{Person}  --> \texttt{Writer}\\ 
    & & \texttt{Person} --> \textbf{\texttt{Politician}} \\
    \midrule
    Wrong Level & \textit{Dolly} is running on the grassland.& 
    \texttt{Horse} --> \underline{\texttt{Mammal}} --> \textbf{\texttt{Animal}}\\
    $71.0\%$ & & \\
    \bottomrule
    \end{tabular}
    \caption{Error examples sampled from zero-shot probing results of \BERTbase on the CiC dataset. \textit{Italics} denote entities. \underline{\texttt{Underlines}} denote model predictions. \textbf{\texttt{Texts in bold}} denote answers.}
    \label{tab:task3_case}
\end{table*}

\begin{figure*}
    \centering
    \includegraphics[width=0.92\linewidth]{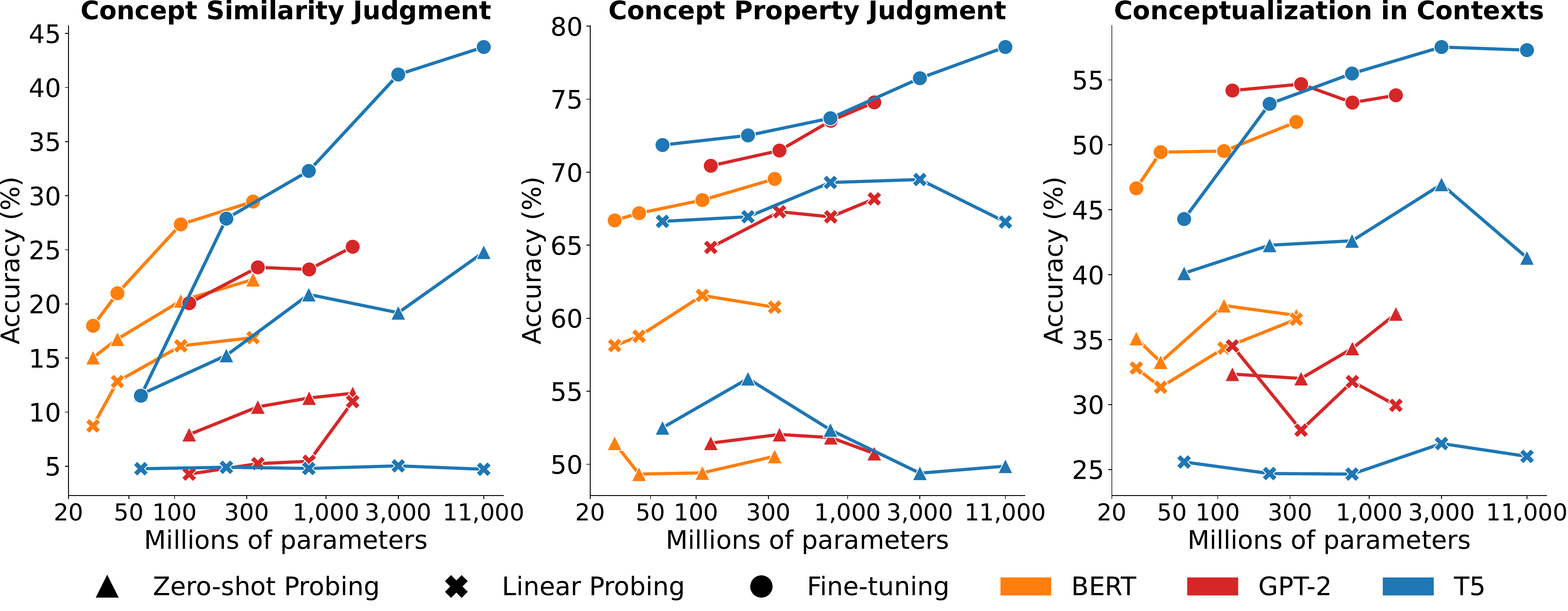}
    \caption{Accuracies (\%) of various PLMs at different scales. The accuracies on CPJ are instance-level.}
    \label{fig:model_scale}
\end{figure*}

\paragraph{Understanding hierarchy is more difficult than disambiguation.}

In \cref{tab:task3_case}, we analyze the two error types on CiC task. \textit{Disambiguation} indicates the PLM selects a wrong concept chain for the given entity and \textit{Wrong Level} indicates the PLM selects a wrong-level concept in the correct chain. In the analysis, we only consider entities with more than one concept chain. 
The \textit{Wrong Level} errors take up the majority, which shows that understanding concept hierarchy is more difficult than disambiguation for PLMs and how to teach the PLMs to understand it is essential.

\subsection{Analysis on Model Scale}
\label{sec:analyze_scale}

Inspired by recent advances showing the superior advantages of large-scale models~\citep{kaplan2020scaling,DBLP:conf/emnlp/LesterAC21}, we explore how the model scale influences PLMs' conceptual knowledge. We investigate the family of three representative PLMs: BERT, GPT-2 and T5. Since fine-tuning extremely-large PLMs is too computationally expensive, for models with more than $2.5$ billion parameters, we instead adopt BitFit~\citep{zaken2021bitfit}, which can achieve similar performance to fine-tuning~\citep{he2021towards} but requires much less computation. The results are shown in \cref{fig:model_scale}, and we have following observations: 
(1) Larger-scale PLMs generally achieve better performance on all the probing tasks, which suggests that increasing model scale can store more conceptual knowledge. However, the improvements brought by increasing model scale are generally marginal, especially on CiC task, and the improvements in zero-shot probing and linear probing results are not so obvious like in fine-tuning, which poses a question that whether the fine-tuning improvements come from the intrinsic knowledge of PLMs.
(2) The fine-tuning accuracies of \Txxl with $11$ billion parameters, are still well below ordinary people, which demonstrates that acquiring conceptual knowledge is quite challenging for existing pre-training methods, which encourages further efforts on building concept-aware PLMs.

\section{Related Work}

\paragraph{Knowledge Probing}

To understand the success of PLMs, extensive works explore to know what PLMs know, and find PLMs have strong linguistic knowledge~\citep{ DBLP:conf/naacl/Liu0BPS19, hewitt-manning-2019-structural, tenney2019you, DBLP:conf/emnlp/VulicPLGK20}. Moreover, it has been shown that PLMs have a certain world knowledge, which is typically stored in world knowledge bases, such as the knowledge about entities~\citep{broscheit-2019-investigating,tenney-etal-2019-bert} and their relationships~\citep{DBLP:conf/emnlp/PetroniRRLBWM19, DBLP:conf/emnlp/RobertsRS20, DBLP:journals/tacl/JiangXAN20, DBLP:conf/aaai/BouraouiCS20, DBLP:conf/naacl/ZhongFC21}. However, these explorations are limited in the scope of factual knowledge, ignoring the conceptual knowledge, which is essential for both knowledge bases~\citep{wu2012probase,ji2019microsoft} and intelligence~\citep{carey1991knowledge,Collins2014KnowledgeIP}. Hence we explore the conceptual knowledge probing in this paper.

\paragraph{Conceptual Knowledge in PLMs}

Previous works also explore the \textit{concept} in PLMs~\citep{DBLP:conf/emnlp/MichaelBT20, DBLP:journals/tacl/TalmorEGB20, DBLP:conf/acl/AspillagaMS21, dalvi2021discovering}, which study principally similar topics with us. However, the \textit{concept} they refer to is essentially \textit{word sense}.  They focus on whether PLMs discover the word senses and recognize their hierarchical relations. While in this work, we study the concepts defined in knowledge bases to abstract real-world entities, which support broader applications~\citep{lv-etal-2018-differentiating,zhou-etal-2021-kacc,zeng2021encoding}, and probe knowledge about conceptual similarity and properties of concepts
as well as PLMs' conceptualization ability.

\section{Conclusion and Future Work}
In this paper, we systematically analyze the conceptual knowledge in existing PLMs by constructing a high-quality conceptual knowledge probing benchmark (\ourdataset). Extensive experiments show that existing PLMs have a certain conceptual knowledge, but are significantly worse than humans, even with billions of parameters. We further find that PLMs fail in distinguishing fine-grained concepts and understanding concept hierarchy, and suffer from conceptual hallucination caused by word occurrence and out-of-context bias. In the future, inspired by works infusing factual knowledge, we will try to develop conceptual knowledgeable PLMs by exploring concept-aware pre-training objectives and knowledge-enhanced architectures.

\section*{Limitations}

In the section, we discuss the limitations of this work: (1) \textbf{COPEN benchmark.} COPEN only involves English corpora, which limits the use of the benchmark to PLMs pre-trained on other languages. In the future, we will consider more languages and construct multilingual \ourdataset. (2) \textbf{Large PLMs.} We do not experiment on very large PLMs, such as GPT-3~\citep{gpt3} and PaLM~\citep{PaLM}, due to our limited access to them. We conduct experiments on \Txxl with 11 billion parameters instead. Experimental results demonstrate that acquiring conceptual knowledge is quite challenging for existing pre-training methods, which urges concept-aware pre-training objectives and model architectures. (3) \textbf{Environmental impact. } In this paper, we conduct a lot of experiments with various PLMs, some of which even contain several billions of parameters. It consumes large amounts of energy and causes large amounts of carbon dioxide emissions, which incurs negative influence to our environment~\citep{DBLP:conf/acl/StrubellGM19}. But the experiments are necessary for drawing faithful and comprehensive conclusions. We hope our findings could facilitate further research on more powerful PLMs with fewer parameters.

\section*{Ethical Considerations}
We discuss the ethical considerations and broader
impact of this work in this section: (1)
\textbf{Intellectual property}. 
The Wikipedia, Simple Wikipedia corpora, and Wikidata are obtained from the Wikimedia dump\footnote{\url{https://dumps.wikimedia.org/}},
which is shared under the CC BY-SA 3.0 license\footnote{\url{https://creativecommons.org/licenses/by-sa/3.0/}}. 
The DBpedia\footnote{\url{www.dbpedia.org}} is shared under the CC BY-SA 3.0 license and GNU Free Documentation License\footnote{\url{https://www.gnu.org/licenses/fdl-1.3.html}}.
The GenericsKB corpus\footnote{\url{https://allenai.org/data/genericskb}} is shared under the CC BY 4.0 license\footnote{\url{https://creativecommons.org/licenses/by/4.0/}}. These are all public and established resources, which are intended to support broad artificial intelligence and NLP research. We believe these resources are well desensitized and anonymized.
(2) \textbf{Data annotation.} We invite $19$ annotators without background of expertise to annotate our datasets and produce human performance. They are all employed by commercial data production companies. The invited annotators are fairly paid according to agreed working hours and prices. The annotators are all informed about how the data will be processed, used, and released, and this is confirmed in the data production contract. 
(3) \textbf{Intended use.} COPEN is a high-quality benchmark used for evaluating conceptual knowledge in PLMs and developing concept-knowledgeable PLMs. Researchers can use COPEN to assess new concept-aware objectives and conceptual-knowledge-enhanced architectures. 
(4) \textbf{Misuse risks.} Considering \ourdataset is built on top of a limited scope of natural texts and the probing methods are inevitably influenced by some spurious correlations, a good enough performance on \ourdataset cannot fully guarantee that the developed methods really understand concepts and shall not be used to support relevant commercial and political claims.
(5) \textbf{Potential risks control.} The texts in \ourdataset are from public data and do not involve private information, sensitive topics and social issues. The three tasks in \ourdataset also do not involve sensitive topics or social issues. We manually check some randomly sampled instances in \ourdataset and find no 
sensitive information or other risky issues. Hence we believe that \ourdataset does not create additional risks.

\section*{Acknowledgements}
This work is supported by the Key-Area Research and Development Program of Guangdong Province (2019B010153002), the Institute for Guo Qiang, Tsinghua University (2019GQB0003), and Huawei Noah's Ark Lab.
The authors thank all the anonymous reviewers for their detailed and valuable comments and suggestions. The authors also thank all the annotators for their substantial efforts in the annotation process.

\bibliography{anthology,custom}

\begin{thebibliography}{65}
\expandafter\ifx\csname natexlab\endcsname\relax\def\natexlab#1{#1}\fi

\bibitem[{Alain and Bengio(2017)}]{DBLP:conf/iclr/AlainB17}
Guillaume Alain and Yoshua Bengio. 2017.
\newblock \href {https://openreview.net/forum?id=HJ4-rAVtl} {Understanding
  intermediate layers using linear classifier probes}.
\newblock In \emph{Proceedings of ICLR}.

\bibitem[{Antoniou and Van~Harmelen(2004)}]{DBLP:books/daglib/0036180}
Grigoris Antoniou and Frank Van~Harmelen. 2004.
\newblock \href
  {http://people.mpi-inf.mpg.de/~dstepano/KRSW/literature/SWPrimer.pdf}
  {\emph{A semantic web primer}}.
\newblock MIT press.

\bibitem[{Aspillaga et~al.(2021)Aspillaga, Mendoza, and
  Soto}]{DBLP:conf/acl/AspillagaMS21}
Carlos Aspillaga, Marcelo Mendoza, and Alvaro Soto. 2021.
\newblock \href {https://doi.org/10.18653/v1/2021.findings-acl.263} {Inspecting
  the concept knowledge graph encoded by modern language models}.
\newblock In \emph{Findings of ACL-IJCNLP}, pages 2984--3000.

\bibitem[{Auer et~al.(2007)Auer, Bizer, Kobilarov, Lehmann, Cyganiak, and
  Ives}]{auer2007dbpedia}
S{\"o}ren Auer, Christian Bizer, Georgi Kobilarov, Jens Lehmann, Richard
  Cyganiak, and Zachary Ives. 2007.
\newblock \href
  {https://link.springer.com/chapter/10.1007/978-3-540-76298-0_52} {D{B}pedia:
  A nucleus for a web of open data}.
\newblock In \emph{The semantic web}, pages 722--735. Springer.

\bibitem[{Bhakthavatsalam et~al.(2020)Bhakthavatsalam, Anastasiades, and
  Clark}]{DBLP:journals/corr/abs-2005-00660}
Sumithra Bhakthavatsalam, Chloe Anastasiades, and Peter Clark. 2020.
\newblock \href {http://arxiv.org/abs/2005.00660} {Generics{KB}: {A} knowledge
  base of generic statements}.
\newblock \emph{CoRR}, abs/2005.00660.

\bibitem[{Black et~al.(2021)Black, Gao, Wang, Leahy, and Biderman}]{gpt-neo}
Sid Black, Leo Gao, Phil Wang, Connor Leahy, and Stella Biderman. 2021.
\newblock \href {https://doi.org/10.5281/zenodo.5297715} {{GPT-Neo: Large Scale
  Autoregressive Language Modeling with Mesh-Tensorflow}}.
\newblock {Z}enodo.

\bibitem[{Bouraoui et~al.(2020)Bouraoui, Camacho{-}Collados, and
  Schockaert}]{DBLP:conf/aaai/BouraouiCS20}
Zied Bouraoui, Jos{\'{e}} Camacho{-}Collados, and Steven Schockaert. 2020.
\newblock \href {https://aaai.org/ojs/index.php/AAAI/article/view/6242}
  {Inducing relational knowledge from {BERT}}.
\newblock In \emph{Proceedings of AAAI-IAAI-EAAI}, pages 7456--7463.

\bibitem[{Broscheit(2019)}]{broscheit-2019-investigating}
Samuel Broscheit. 2019.
\newblock \href {https://doi.org/10.18653/v1/K19-1063} {Investigating entity
  knowledge in {BERT} with simple neural end-to-end entity linking}.
\newblock In \emph{Proceedings of CoNLL}, pages 677--685.

\bibitem[{Brown et~al.(2020)Brown, Mann, Ryder, Subbiah, Kaplan, Dhariwal,
  Neelakantan, Shyam, Sastry, Askell, Agarwal, Herbert{-}Voss, Krueger,
  Henighan, Child, Ramesh, Ziegler, Wu, Winter, Hesse, Chen, Sigler, Litwin,
  Gray, Chess, Clark, Berner, McCandlish, Radford, Sutskever, and
  Amodei}]{gpt3}
Tom~B. Brown, Benjamin Mann, Nick Ryder, Melanie Subbiah, Jared Kaplan,
  Prafulla Dhariwal, Arvind Neelakantan, Pranav Shyam, Girish Sastry, Amanda
  Askell, Sandhini Agarwal, Ariel Herbert{-}Voss, Gretchen Krueger, Tom
  Henighan, Rewon Child, Aditya Ramesh, Daniel~M. Ziegler, Jeffrey Wu, Clemens
  Winter, Christopher Hesse, Mark Chen, Eric Sigler, Mateusz Litwin, Scott
  Gray, Benjamin Chess, Jack Clark, Christopher Berner, Sam McCandlish, Alec
  Radford, Ilya Sutskever, and Dario Amodei. 2020.
\newblock \href
  {https://proceedings.neurips.cc/paper/2020/hash/1457c0d6bfcb4967418bfb8ac142f64a-Abstract.html}
  {Language models are few-shot learners}.
\newblock In \emph{Proceedings of NeurIPS}, pages 1877--1901.

\bibitem[{Cao et~al.(2021)Cao, Lin, Han, Sun, Yan, Liao, Xue, and
  Xu}]{DBLP:conf/acl/CaoLHSYLXX20}
Boxi Cao, Hongyu Lin, Xianpei Han, Le~Sun, Lingyong Yan, Meng Liao, Tong Xue,
  and Jin Xu. 2021.
\newblock \href {https://doi.org/10.18653/v1/2021.acl-long.146} {Knowledgeable
  or educated guess? {R}evisiting language models as knowledge bases}.
\newblock In \emph{Proceedings of ACL-IJCNLP}, pages 1860--1874.

\bibitem[{Carey(1991)}]{carey1991knowledge}
Susan Carey. 1991.
\newblock \href
  {https://pcl.sitehost.iu.edu/rgoldsto/courses/concepts/carey.pdf} {Knowledge
  acquisition: Enrichment or conceptual change}.
\newblock \emph{The epigenesis of mind: Essays on biology and cognition}, pages
  257--291.

\bibitem[{Chowdhery et~al.(2022)Chowdhery, Narang, Devlin, Bosma, Mishra,
  Roberts, Barham, Chung, Sutton, Gehrmann, Schuh, Shi, Tsvyashchenko, Maynez,
  Rao, Barnes, Tay, Shazeer, Prabhakaran, Reif, Du, Hutchinson, Pope, Bradbury,
  Austin, Isard, Gur{-}Ari, Yin, Duke, Levskaya, Ghemawat, Dev, Michalewski,
  Garcia, Misra, Robinson, Fedus, Zhou, Ippolito, Luan, Lim, Zoph, Spiridonov,
  Sepassi, Dohan, Agrawal, Omernick, Dai, Pillai, Pellat, Lewkowycz, Moreira,
  Child, Polozov, Lee, Zhou, Wang, Saeta, Diaz, Firat, Catasta, Wei,
  Meier{-}Hellstern, Eck, Dean, Petrov, and Fiedel}]{PaLM}
Aakanksha Chowdhery, Sharan Narang, Jacob Devlin, Maarten Bosma, Gaurav Mishra,
  Adam Roberts, Paul Barham, Hyung~Won Chung, Charles Sutton, Sebastian
  Gehrmann, Parker Schuh, Kensen Shi, Sasha Tsvyashchenko, Joshua Maynez,
  Abhishek Rao, Parker Barnes, Yi~Tay, Noam Shazeer, Vinodkumar Prabhakaran,
  Emily Reif, Nan Du, Ben Hutchinson, Reiner Pope, James Bradbury, Jacob
  Austin, Michael Isard, Guy Gur{-}Ari, Pengcheng Yin, Toju Duke, Anselm
  Levskaya, Sanjay Ghemawat, Sunipa Dev, Henryk Michalewski, Xavier Garcia,
  Vedant Misra, Kevin Robinson, Liam Fedus, Denny Zhou, Daphne Ippolito, David
  Luan, Hyeontaek Lim, Barret Zoph, Alexander Spiridonov, Ryan Sepassi, David
  Dohan, Shivani Agrawal, Mark Omernick, Andrew~M. Dai,
  Thanumalayan~Sankaranarayana Pillai, Marie Pellat, Aitor Lewkowycz, Erica
  Moreira, Rewon Child, Oleksandr Polozov, Katherine Lee, Zongwei Zhou, Xuezhi
  Wang, Brennan Saeta, Mark Diaz, Orhan Firat, Michele Catasta, Jason Wei,
  Kathy Meier{-}Hellstern, Douglas Eck, Jeff Dean, Slav Petrov, and Noah
  Fiedel. 2022.
\newblock \href {https://doi.org/10.48550/arXiv.2204.02311} {Pa{LM}: Scaling
  language modeling with pathways}.
\newblock \emph{CoRR}, abs/2204.02311.

\bibitem[{Collins and Olson(2014)}]{Collins2014KnowledgeIP}
Jessica~A. Collins and Ingrid~R. Olson. 2014.
\newblock \href {https://link.springer.com/article/10.3758/s13423-013-0564-3}
  {Knowledge is power: How conceptual knowledge transforms visual cognition}.
\newblock \emph{Psychonomic Bulletin \& Review}, 21:843--860.

\bibitem[{Cruse(1986)}]{cruse1986lexical}
David~Alan Cruse. 1986.
\newblock \emph{Lexical semantics}.
\newblock Cambridge university press.

\bibitem[{Dalvi et~al.(2021)Dalvi, Khan, Alam, Durrani, Xu, and
  Sajjad}]{dalvi2021discovering}
Fahim Dalvi, Abdul~Rafae Khan, Firoj Alam, Nadir Durrani, Jia Xu, and Hassan
  Sajjad. 2021.
\newblock \href {https://openreview.net/forum?id=POTMtpYI1xH} {Discovering
  latent concepts learned in {BERT}}.
\newblock In \emph{Proceedings of ICLR}.

\bibitem[{Decker et~al.(2000)Decker, Melnik, van Harmelen, Fensel, Klein,
  Broekstra, Erdmann, and Horrocks}]{DBLP:journals/internet/DeckerMHFKBEH00}
Stefan Decker, Sergey Melnik, Frank van Harmelen, Dieter Fensel, Michel C.~A.
  Klein, Jeen Broekstra, Michael Erdmann, and Ian Horrocks. 2000.
\newblock \href {https://doi.org/10.1109/4236.877487} {The semantic web: The
  roles of {XML} and {RDF}}.
\newblock \emph{{IEEE} Internet Comput.}, 4(5):63--74.

\bibitem[{Devlin et~al.(2019)Devlin, Chang, Lee, and
  Toutanova}]{DBLP:conf/naacl/DevlinCLT19}
Jacob Devlin, Ming-Wei Chang, Kenton Lee, and Kristina Toutanova. 2019.
\newblock \href {https://doi.org/10.18653/v1/N19-1423} {{BERT}: Pre-training of
  deep bidirectional transformers for language understanding}.
\newblock In \emph{Proceedings of NAACL-HLT}, pages 4171--4186.

\bibitem[{Guha et~al.(2016)Guha, Brickley, and Macbeth}]{guha2016schema}
Ramanathan~V Guha, Dan Brickley, and Steve Macbeth. 2016.
\newblock \href {https://dl.acm.org/doi/fullHtml/10.1145/2844544} {Schema. org:
  {E}volution of structured data on the web}.
\newblock \emph{Communications of the ACM}, 59(2):44--51.

\bibitem[{Han et~al.(2021)Han, Zhang, Ding, Gu, Liu, Huo, Qiu, Zhang, Han,
  Huang et~al.}]{han2021pre}
Xu~Han, Zhengyan Zhang, Ning Ding, Yuxian Gu, Xiao Liu, Yuqi Huo, Jiezhong Qiu,
  Liang Zhang, Wentao Han, Minlie Huang, et~al. 2021.
\newblock \href {https://doi.org/10.1016/j.aiopen.2021.08.002.} {Pre-trained
  models: Past, present and future}.
\newblock \emph{Proceedings of AI Open}.

\bibitem[{He et~al.(2021)He, Zhou, Ma, Berg-Kirkpatrick, and
  Neubig}]{he2021towards}
Junxian He, Chunting Zhou, Xuezhe Ma, Taylor Berg-Kirkpatrick, and Graham
  Neubig. 2021.
\newblock \href {https://openreview.net/pdf?id=0RDcd5Axok} {Towards a unified
  view of parameter-efficient transfer learning}.
\newblock \emph{arXiv preprint arXiv:2110.04366}.

\bibitem[{Hewitt and Manning(2019)}]{hewitt-manning-2019-structural}
John Hewitt and Christopher~D. Manning. 2019.
\newblock \href {https://doi.org/10.18653/v1/N19-1419} {{A} structural probe
  for finding syntax in word representations}.
\newblock In \emph{Proceedings of NAACL-HLT}, pages 4129--4138.

\bibitem[{Hill et~al.(2015)Hill, Reichart, and Korhonen}]{simlex-999}
Felix Hill, Roi Reichart, and Anna Korhonen. 2015.
\newblock \href {https://doi.org/10.1162/COLI\_a\_00237} {Simlex-999:
  Evaluating semantic models with (genuine) similarity estimation}.
\newblock \emph{Comput. Linguistics}, 41(4):665--695.

\bibitem[{Ji et~al.(2019)Ji, Wang, Shi, Zhang, Wang, and Yan}]{ji2019microsoft}
Lei Ji, Yujing Wang, Botian Shi, Dawei Zhang, Zhongyuan Wang, and Jun Yan.
  2019.
\newblock \href
  {https://direct.mit.edu/dint/article/1/3/238/9983/Microsoft-Concept-Graph-Mining-Semantic-Concepts}
  {Microsoft concept graph: Mining semantic concepts for short text
  understanding}.
\newblock \emph{Data Intelligence}, 1(3):238--270.

\bibitem[{Ji et~al.(2022)Ji, Lee, Frieske, Yu, Su, Xu, Ishii, Bang, Madotto,
  and Fung}]{DBLP:journals/corr/abs-2202-03629}
Ziwei Ji, Nayeon Lee, Rita Frieske, Tiezheng Yu, Dan Su, Yan Xu, Etsuko Ishii,
  Yejin Bang, Andrea Madotto, and Pascale Fung. 2022.
\newblock \href {http://arxiv.org/abs/2202.03629} {Survey of hallucination in
  natural language generation}.
\newblock \emph{CoRR}, abs/2202.03629.

\bibitem[{Jiang et~al.(2020)Jiang, Xu, Araki, and
  Neubig}]{DBLP:journals/tacl/JiangXAN20}
Zhengbao Jiang, Frank~F. Xu, Jun Araki, and Graham Neubig. 2020.
\newblock \href {https://transacl.org/ojs/index.php/tacl/article/view/1983}
  {How can we know what language models know}.
\newblock \emph{Trans. Assoc. Comput. Linguistics}, 8:423--438.

\bibitem[{Kaplan et~al.(2020)Kaplan, McCandlish, Henighan, Brown, Chess, Child,
  Gray, Radford, Wu, and Amodei}]{kaplan2020scaling}
Jared Kaplan, Sam McCandlish, Tom Henighan, Tom~B Brown, Benjamin Chess, Rewon
  Child, Scott Gray, Alec Radford, Jeffrey Wu, and Dario Amodei. 2020.
\newblock \href {https://arxiv.org/abs/2001.08361} {Scaling laws for neural
  language models}.
\newblock \emph{arXiv preprint arXiv:2001.08361}.

\bibitem[{Lastra{-}D{\'{\i}}az et~al.(2019)Lastra{-}D{\'{\i}}az, Goikoetxea,
  Taieb, Garc{\'{\i}}a{-}Serrano, Aouicha, and Agirre}]{word-embedding-compare}
Juan~J. Lastra{-}D{\'{\i}}az, Josu Goikoetxea, Mohamed Ali~Hadj Taieb, Ana
  Garc{\'{\i}}a{-}Serrano, Mohamed~Ben Aouicha, and Eneko Agirre. 2019.
\newblock \href {https://doi.org/10.1016/j.engappai.2019.07.010} {A
  reproducible survey on word embeddings and ontology-based methods for word
  similarity: Linear combinations outperform the state of the art}.
\newblock \emph{Eng. Appl. Artif. Intell.}, 85:645--665.

\bibitem[{Lester et~al.(2021)Lester, Al{-}Rfou, and
  Constant}]{DBLP:conf/emnlp/LesterAC21}
Brian Lester, Rami Al{-}Rfou, and Noah Constant. 2021.
\newblock \href {https://doi.org/10.18653/v1/2021.emnlp-main.243} {The power of
  scale for parameter-efficient prompt tuning}.
\newblock In \emph{Proceedings of EMNLP}, pages 3045--3059.

\bibitem[{Lewis et~al.(2020)Lewis, Liu, Goyal, Ghazvininejad, Mohamed, Levy,
  Stoyanov, and Zettlemoyer}]{DBLP:conf/acl/LewisLGGMLSZ20}
Mike Lewis, Yinhan Liu, Naman Goyal, Marjan Ghazvininejad, Abdelrahman Mohamed,
  Omer Levy, Veselin Stoyanov, and Luke Zettlemoyer. 2020.
\newblock \href {https://doi.org/10.18653/v1/2020.acl-main.703} {{BART}:
  Denoising sequence-to-sequence pre-training for natural language generation,
  translation, and comprehension}.
\newblock In \emph{Proceedings of ACL}, pages 7871--7880.

\bibitem[{Liu et~al.(2019{\natexlab{a}})Liu, Gardner, Belinkov, Peters, and
  Smith}]{DBLP:conf/naacl/Liu0BPS19}
Nelson~F. Liu, Matt Gardner, Yonatan Belinkov, Matthew~E. Peters, and Noah~A.
  Smith. 2019{\natexlab{a}}.
\newblock \href {https://doi.org/10.18653/v1/N19-1112} {Linguistic knowledge
  and transferability of contextual representations}.
\newblock In \emph{Proceedings of NAACL-HLT}, pages 1073--1094.

\bibitem[{Liu et~al.(2021{\natexlab{a}})Liu, Yuan, Fu, Jiang, Hayashi, and
  Neubig}]{DBLP:journals/corr/abs-2107-13586}
Pengfei Liu, Weizhe Yuan, Jinlan Fu, Zhengbao Jiang, Hiroaki Hayashi, and
  Graham Neubig. 2021{\natexlab{a}}.
\newblock \href {http://arxiv.org/abs/2107.13586} {Pre-train, prompt, and
  predict: {A} systematic survey of prompting methods in natural language
  processing}.
\newblock \emph{CoRR}, abs/2107.13586.

\bibitem[{Liu et~al.(2021{\natexlab{b}})Liu, Zheng, Du, Ding, Qian, Yang, and
  Tang}]{p-tuning}
Xiao Liu, Yanan Zheng, Zhengxiao Du, Ming Ding, Yujie Qian, Zhilin Yang, and
  Jie Tang. 2021{\natexlab{b}}.
\newblock \href {http://arxiv.org/abs/2103.10385} {{GPT} understands, too}.
\newblock \emph{CoRR}, abs/2103.10385.

\bibitem[{Liu et~al.(2019{\natexlab{b}})Liu, Ott, Goyal, Du, Joshi, Chen, Levy,
  Lewis, Zettlemoyer, and Stoyanov}]{DBLP:journals/corr/abs-1907-11692}
Yinhan Liu, Myle Ott, Naman Goyal, Jingfei Du, Mandar Joshi, Danqi Chen, Omer
  Levy, Mike Lewis, Luke Zettlemoyer, and Veselin Stoyanov. 2019{\natexlab{b}}.
\newblock \href {http://arxiv.org/abs/1907.11692} {Ro{BERT}a: {A} robustly
  optimized {BERT} pretraining approach}.
\newblock \emph{CoRR}, abs/1907.11692.

\bibitem[{Lv et~al.(2018)Lv, Hou, Li, and Liu}]{lv-etal-2018-differentiating}
Xin Lv, Lei Hou, Juanzi Li, and Zhiyuan Liu. 2018.
\newblock \href {https://doi.org/10.18653/v1/D18-1222} {Differentiating
  concepts and instances for knowledge graph embedding}.
\newblock In \emph{Proceedings of EMNLP}, pages 1971--1979.

\bibitem[{McGuinness et~al.(2004)McGuinness, Van~Harmelen
  et~al.}]{mcguinness2004owl}
Deborah~L McGuinness, Frank Van~Harmelen, et~al. 2004.
\newblock \href {https://www.w3.org/TR/owl-features/} {Owl web ontology
  language overview}.
\newblock \emph{W3C recommendation}, 10(10):2004.

\bibitem[{Mchaput(2016)}]{mchaput_2016}
Mchaput. 2016.
\newblock \href {https://github.com/mchaput/whoosh} {Mchaput/whoosh:
  Pure-python full-text search library}.
\newblock GitHub.

\bibitem[{Michael et~al.(2020)Michael, Botha, and
  Tenney}]{DBLP:conf/emnlp/MichaelBT20}
Julian Michael, Jan~A. Botha, and Ian Tenney. 2020.
\newblock \href {https://doi.org/10.18653/v1/2020.emnlp-main.552} {Asking
  without telling: Exploring latent ontologies in contextual representations}.
\newblock In \emph{Proceedings of EMNLP}, pages 6792--6812.

\bibitem[{Murphy(2004)}]{murphy2004big}
Gregory Murphy. 2004.
\newblock \href
  {http://dspace.stellamariscollege.edu.in:8080/xmlui/bitstream/handle/123456789/1778/The\%20big\%20book\%20of\%20concepts.pdf?sequence=1&isAllowed=y}
  {\emph{The big book of concepts}}.
\newblock MIT press.

\bibitem[{Pennington et~al.(2014)Pennington, Socher, and Manning}]{glove}
Jeffrey Pennington, Richard Socher, and Christopher~D. Manning. 2014.
\newblock \href {https://doi.org/10.3115/v1/d14-1162} {Glove: Global vectors
  for word representation}.
\newblock In \emph{Proceedings of EMNLP}, pages 1532--1543.

\bibitem[{Petroni et~al.(2019)Petroni, Rockt{\"a}schel, Riedel, Lewis, Bakhtin,
  Wu, and Miller}]{DBLP:conf/emnlp/PetroniRRLBWM19}
Fabio Petroni, Tim Rockt{\"a}schel, Sebastian Riedel, Patrick Lewis, Anton
  Bakhtin, Yuxiang Wu, and Alexander Miller. 2019.
\newblock \href {https://doi.org/10.18653/v1/D19-1250} {Language models as
  knowledge bases?}
\newblock In \emph{Proceedings of EMNLP-IJCNLP}, pages 2463--2473.

\bibitem[{Prechelt(1996)}]{DBLP:conf/nips/Prechelt96}
Lutz Prechelt. 1996.
\newblock \href {https://doi.org/10.1007/3-540-49430-8\_3} {Early stopping-but
  when?}
\newblock In Genevieve~B. Orr and Klaus{-}Robert M{\"{u}}ller, editors,
  \emph{Neural Networks: Tricks of the Trade}, volume 1524 of \emph{Lecture
  Notes in Computer Science}, pages 55--69. Springer.

\bibitem[{Qiu et~al.(2020)Qiu, Sun, Xu, Shao, Dai, and Huang}]{qiu2020pre}
Xipeng Qiu, Tianxiang Sun, Yige Xu, Yunfan Shao, Ning Dai, and Xuanjing Huang.
  2020.
\newblock \href {https://link.springer.com/article/10.1007/s11431-020-1647-3}
  {Pre-trained models for natural language processing: A survey}.
\newblock \emph{Science China Technological Sciences}, 63(10):1872--1897.

\bibitem[{Radford et~al.(2019)Radford, Wu, Child, Luan, Amodei, Sutskever
  et~al.}]{radford2019language}
Alec Radford, Jeffrey Wu, Rewon Child, David Luan, Dario Amodei, Ilya
  Sutskever, et~al. 2019.
\newblock \href {http://www.persagen.com/files/misc/radford2019language.pdf}
  {Language models are unsupervised multitask learners}.
\newblock \emph{OpenAI blog}, 1(8):9.

\bibitem[{Raffel et~al.(2020)Raffel, Shazeer, Roberts, Lee, Narang, Matena,
  Zhou, Li, and Liu}]{DBLP:journals/jmlr/RaffelSRLNMZLL20}
Colin Raffel, Noam Shazeer, Adam Roberts, Katherine Lee, Sharan Narang, Michael
  Matena, Yanqi Zhou, Wei Li, and Peter~J. Liu. 2020.
\newblock \href {http://jmlr.org/papers/v21/20-074.html} {Exploring the limits
  of transfer learning with a unified text-to-text transformer}.
\newblock \emph{J. Mach. Learn. Res.}, 21:140:1--140:67.

\bibitem[{Reiter(2018)}]{hallucination-in-neural-nlg}
Ehud Reiter. 2018.
\newblock \href
  {https://ehudreiter.com/2018/11/12/hallucination-in-neural-nlg/}
  {Hallucination in {N}eural {NLG}}.
\newblock Ehud Reiter's Blog.

\bibitem[{Roberts et~al.(2020)Roberts, Raffel, and
  Shazeer}]{DBLP:conf/emnlp/RobertsRS20}
Adam Roberts, Colin Raffel, and Noam Shazeer. 2020.
\newblock \href {https://doi.org/10.18653/v1/2020.emnlp-main.437} {How much
  knowledge can you pack into the parameters of a language model?}
\newblock In \emph{Proceedings of EMNLP}, pages 5418--5426.

\bibitem[{Robertson et~al.(1995)Robertson, Walker, Jones, Hancock-Beaulieu,
  Gatford et~al.}]{robertson1995okapi}
Stephen~E Robertson, Steve Walker, Susan Jones, Micheline~M Hancock-Beaulieu,
  Mike Gatford, et~al. 1995.
\newblock \href
  {https://www.researchgate.net/profile/Stephen-Robertson-11/publication/254309988_Okapi_at_trec4/links/00b495346c7b7cd38d000000/Okapi-at-trec4.pdf}
  {Okapi at {TREC}-3}.
\newblock \emph{Nist Special Publication Sp}, 109:109.

\bibitem[{Rohrbach et~al.(2018)Rohrbach, Hendricks, Burns, Darrell, and
  Saenko}]{DBLP:conf/emnlp/RohrbachHBDS18}
Anna Rohrbach, Lisa~Anne Hendricks, Kaylee Burns, Trevor Darrell, and Kate
  Saenko. 2018.
\newblock \href {https://doi.org/10.18653/v1/D18-1437} {Object {H}allucination
  in {I}mage {C}aptioning}.
\newblock In \emph{Proceedings of EMNLP}, pages 4035--4045.

\bibitem[{Sowa(1976)}]{sowa1976conceptual}
John~F Sowa. 1976.
\newblock \href {https://dl.acm.org/doi/abs/10.1147/rd.204.0336} {Conceptual
  graphs for a data base interface}.
\newblock \emph{IBM Journal of Research and Development}, 20(4):336--357.

\bibitem[{Strubell et~al.(2019)Strubell, Ganesh, and
  McCallum}]{DBLP:conf/acl/StrubellGM19}
Emma Strubell, Ananya Ganesh, and Andrew McCallum. 2019.
\newblock \href {https://doi.org/10.18653/v1/p19-1355} {Energy and policy
  considerations for deep learning in {NLP}}.
\newblock In \emph{Proceedings of ACL}, pages 3645--3650.

\bibitem[{Suchanek et~al.(2007)Suchanek, Kasneci, and Weikum}]{yago}
Fabian~M. Suchanek, Gjergji Kasneci, and Gerhard Weikum. 2007.
\newblock \href {https://doi.org/10.1145/1242572.1242667} {Yago: a core of
  semantic knowledge}.
\newblock In \emph{Proceedings of WWW}, pages 697--706.

\bibitem[{Talmor et~al.(2020)Talmor, Elazar, Goldberg, and
  Berant}]{DBLP:journals/tacl/TalmorEGB20}
Alon Talmor, Yanai Elazar, Yoav Goldberg, and Jonathan Berant. 2020.
\newblock \href {https://transacl.org/ojs/index.php/tacl/article/view/2041}
  {o{LM}pics - {O}n what {L}anguage {M}odel {P}re-training {C}aptures}.
\newblock \emph{Trans. Assoc. Comput. Linguistics}, 8:743--758.

\bibitem[{Tenney et~al.(2019{\natexlab{a}})Tenney, Das, and
  Pavlick}]{tenney-etal-2019-bert}
Ian Tenney, Dipanjan Das, and Ellie Pavlick. 2019{\natexlab{a}}.
\newblock \href {https://doi.org/10.18653/v1/P19-1452} {{BERT} {R}ediscovers
  the {C}lassical {NLP} {P}ipeline}.
\newblock In \emph{Proceedings of ACL}, pages 4593--4601.

\bibitem[{Tenney et~al.(2019{\natexlab{b}})Tenney, Xia, Chen, Wang, Poliak,
  McCoy, Kim, Durme, Bowman, Das, and Pavlick}]{tenney2019you}
Ian Tenney, Patrick Xia, Berlin Chen, Alex Wang, Adam Poliak, R.~Thomas McCoy,
  Najoung Kim, Benjamin~Van Durme, Samuel~R. Bowman, Dipanjan Das, and Ellie
  Pavlick. 2019{\natexlab{b}}.
\newblock \href {https://openreview.net/forum?id=SJzSgnRcKX} {What do you learn
  from context? {P}robing for sentence structure in contextualized word
  representations}.
\newblock In \emph{Proceedings of ICLR}.

\bibitem[{Vrande{\v{c}}i{\'c}(2012)}]{vrandevcic2012wikidata}
Denny Vrande{\v{c}}i{\'c}. 2012.
\newblock \href {https://dl.acm.org/doi/abs/10.1145/2187980.2188242} {Wikidata:
  A new platform for collaborative data collection}.
\newblock In \emph{Proceedings of WWW}, pages 1063--1064.

\bibitem[{Vrande{\v{c}}i{\'c} and Kr{\"o}tzsch(2014)}]{vrandevcic2014wikidata}
Denny Vrande{\v{c}}i{\'c} and Markus Kr{\"o}tzsch. 2014.
\newblock \href {https://dl.acm.org/doi/fullHtml/10.1145/2629489} {Wikidata:
  {A} free collaborative knowledgebase}.
\newblock \emph{Communications of the ACM}, 57(10):78--85.

\bibitem[{Vuli{\'c} et~al.(2020)Vuli{\'c}, Ponti, Litschko, Glava{\v{s}}, and
  Korhonen}]{DBLP:conf/emnlp/VulicPLGK20}
Ivan Vuli{\'c}, Edoardo~Maria Ponti, Robert Litschko, Goran Glava{\v{s}}, and
  Anna Korhonen. 2020.
\newblock \href {https://doi.org/10.18653/v1/2020.emnlp-main.586} {{P}robing
  {P}retrained {L}anguage {M}odels for {L}exical {S}emantics}.
\newblock In \emph{Proceedings of EMNLP}, pages 7222--7240.

\bibitem[{Waxman and Markow(1995)}]{Waxman1995WordsAI}
Sandra~R. Waxman and Dana Markow. 1995.
\newblock \href
  {https://www.sciencedirect.com/science/article/abs/pii/S001002858571016X}
  {Words as {I}nvitations to {F}orm {C}ategories: {E}vidence from 12- to
  13-{M}onth-{O}ld {I}nfants}.
\newblock \emph{Cognitive Psychology}, 29:257--302.

\bibitem[{Wellsby and Pexman(2014)}]{10.3389/fpsyg.2014.00506}
Michele Wellsby and Penny~M. Pexman. 2014.
\newblock \href {https://doi.org/10.3389/fpsyg.2014.00506} {Developing embodied
  cognition: Insights from children’s concepts and language processing}.
\newblock \emph{Frontiers in Psychology}, 5.

\bibitem[{Wolf et~al.(2020)Wolf, Debut, Sanh, Chaumond, Delangue, Moi, Cistac,
  Rault, Louf, Funtowicz, Davison, Shleifer, von Platen, Ma, Jernite, Plu, Xu,
  Le~Scao, Gugger, Drame, Lhoest, and Rush}]{DBLP:conf/emnlp/WolfDSCDMCRLFDS20}
Thomas Wolf, Lysandre Debut, Victor Sanh, Julien Chaumond, Clement Delangue,
  Anthony Moi, Pierric Cistac, Tim Rault, Remi Louf, Morgan Funtowicz, Joe
  Davison, Sam Shleifer, Patrick von Platen, Clara Ma, Yacine Jernite, Julien
  Plu, Canwen Xu, Teven Le~Scao, Sylvain Gugger, Mariama Drame, Quentin Lhoest,
  and Alexander Rush. 2020.
\newblock \href {https://doi.org/10.18653/v1/2020.emnlp-demos.6} {Transformers:
  State-of-the-art natural language processing}.
\newblock In \emph{Proceedings of EMNLP}, pages 38--45.

\bibitem[{Wu et~al.(2012)Wu, Li, Wang, and Zhu}]{wu2012probase}
Wentao Wu, Hongsong Li, Haixun Wang, and Kenny~Q Zhu. 2012.
\newblock \href {https://dl.acm.org/doi/abs/10.1145/2213836.2213891} {Probase:
  A probabilistic taxonomy for text understanding}.
\newblock In \emph{Proceedings of the 2012 ACM SIGMOD International Conference
  on Management of Data}, pages 481--492.

\bibitem[{Zaken et~al.(2022)Zaken, Goldberg, and Ravfogel}]{zaken2021bitfit}
Elad~Ben Zaken, Yoav Goldberg, and Shauli Ravfogel. 2022.
\newblock \href {https://aclanthology.org/2022.acl-short.1} {Bitfit: Simple
  parameter-efficient fine-tuning for transformer-based masked
  language-models}.
\newblock In \emph{Proceedings of ACL}, pages 1--9.

\bibitem[{Zeng et~al.(2021)Zeng, Li, Qi, Lv, Hou, Peng, Li, and
  Feng}]{zeng2021encoding}
Kaisheng Zeng, Chengjiang Li, Yan Qi, Xin Lv, Lei Hou, Guozheng Peng, Juanzi
  Li, and Ling Feng. 2021.
\newblock \href
  {https://link.springer.com/chapter/10.1007/978-3-030-90888-1_19} {Encoding
  the meaning triangle (object, entity, and concept) as the semantic foundation
  for entity alignment}.
\newblock In \emph{Proceedings of WISE}, pages 227--241.

\bibitem[{Zhong et~al.(2021)Zhong, Friedman, and
  Chen}]{DBLP:conf/naacl/ZhongFC21}
Zexuan Zhong, Dan Friedman, and Danqi Chen. 2021.
\newblock \href {https://doi.org/10.18653/v1/2021.naacl-main.398} {Factual
  probing is [{MASK}]: Learning vs. learning to recall}.
\newblock In \emph{Proceedings of NAACL-HLT}, pages 5017--5033.

\bibitem[{Zhou et~al.(2021)Zhou, Hu, Lv, Yang, Liu, Xu, Jiang, Li, and
  Sun}]{zhou-etal-2021-kacc}
Jie Zhou, Shengding Hu, Xin Lv, Cheng Yang, Zhiyuan Liu, Wei Xu, Jie Jiang,
  Juanzi Li, and Maosong Sun. 2021.
\newblock \href {https://doi.org/10.18653/v1/2021.findings-acl.153} {{KACC}: A
  multi-task benchmark for knowledge abstraction, concretization and
  completion}.
\newblock In \emph{Findings of ACL-IJCNLP}, pages 1751--1763.

\end{thebibliography}
\bibliographystyle{acl_natbib}

\newpage
\clearpage
\appendix
\section*{Appendices}
\begin{table}[h!]
    \centering
    \small
    \begin{tabular}{ll}
    \toprule
    Model & \texttt{model\_name} \\
    \midrule
    \BERTsmall & \texttt{prajjwal1/bert-small} \\
    \BERTmedium & \texttt{prajjwal1/bert-medium} \\
    \BERTbase & \texttt{bert-base-uncased}\\
    \BERTlarge & \texttt{bert-large-uncased} \\
    \Rbase & \texttt{roberta-base} \\
    \GPTbase & \texttt{gpt2} \\
    \GPTmedium & \texttt{gpt2-medium} \\
    \GPTlarge & \texttt{gpt2-large} \\
    \GPTxl & \texttt{gpt2-xl} \\
    \GPTNeobase & \texttt{EleutherAI/gpt-neo-125M} \\
    \BARTbase & \texttt{facebook/bart-base} \\
    \Tsmall & \texttt{t5-small} \\
    \Tbase & \texttt{t5-base} \\
    \Tlarge & \texttt{t5-large} \\
    \Txl & \texttt{t5-3b} \\
    \Txxl & \texttt{t5-11b} \\
    \bottomrule
    \end{tabular}
    \caption{The corresponding \texttt{model\_name}s in Transformers library~\citep{DBLP:conf/emnlp/WolfDSCDMCRLFDS20} for different PLMs.}
    \label{tab:appendix_modelname}
\end{table}
\section{Implementation Details}
\label{sec:appendix_experiment}
We use the implementation code and pre-trained parameters of PLMs released in HuggingFace Transformers library~\citep{DBLP:conf/emnlp/WolfDSCDMCRLFDS20} to run our experiments. 
The \texttt{model\_name}s we used in Transformers for different PLMs are shown in \cref{tab:appendix_modelname}. We run experiments for large models (\Txl, and \Txxl) on NVIDIA V100 GPUs, which approximately consumes 160 GPU hours,
and the other PLMs on Nvidia GEFORCE RTX 3090 GPUs, which 
consumes about 300 GPU hours. 
We will introduce the implementation details for zero-shot probing (\cref{sec:appendix_probing}), linear probing (\cref{sec:appendix_linearprobing}), and fine-tuning (\cref{sec:appendix_ft}).

\subsection{Zero-Shot Probing}
\label{sec:appendix_probing}
As mentioned in \cref{sec:probing_method}, we take different text parts of the prompts into scoring calculation.
\cref{tab:appendix_textspan} shows the text parts used by various PLMs to score prompts on the three datasets.

\subsection{Linear Probing}
\label{sec:appendix_linearprobing}
We use the final outputs of specific tokens as the features extracted by PLMs: \texttt{[CLS]} for BERT; \texttt{<s>} for RoBERTa;
the last token for GPT-2, GPT-Neo, and BART; the first token for T5.
We then tune a lightweight linear classifier on the fixed features for BERT, RoBERTa, GPT-2, GPT-Neo, BART and tune the final vocabulary classification head for T5. 
Moreover, we reformulate the original instances into the text-to-text format for T5, and the input and output formats are shown in Table~\ref{tab:appendix_t5}.

\paragraph{Hyperparameters}
We set the learning rate as $1 \times 10^{-3}$ and apply early stopping~\citep{DBLP:conf/nips/Prechelt96} on the accuracy on the development dataset with a patience of 20 epochs. We keep the other hyperparameters the same as in \cref{tab:appendix_hyper}.

\subsection{Fine-Tuning}
\label{sec:appendix_ft}
We follow the fine-tuning methods in original papers to fine-tune BERT~\citep{DBLP:conf/naacl/DevlinCLT19}, RoBERTa~\citep{DBLP:journals/corr/abs-1907-11692}, GPT-2~\citep{radford2019language}, GPT-Neo~\citep{gpt-neo}, and BART~\citep{DBLP:conf/acl/LewisLGGMLSZ20}.  As in \cref{sec:appendix_linearprobing}, we reformulate the original instances into the text-to-text format for T5~\citep{DBLP:journals/jmlr/RaffelSRLNMZLL20}, and the input and output formats are shown in Table~\ref{tab:appendix_t5}.

\paragraph{Hyperparameters}
We follow the hyperparameters mostly used in previous literature. The hyperparameters are shown in Table~\ref{tab:appendix_hyper}. 
And we apply early stopping~\citep{DBLP:conf/nips/Prechelt96} on the accuracy on the development dataset. 

\paragraph{Parameter-efficient Tuning for Big Models}
Due to the limits of computation, we consider the parameter-efficient tuning for models with more than $2.5$ billion parameters (\Txl and \Txxl). Previous works~\citep{he2021towards} have proven that
parameter-efficient tuning methods can save GPU memory, accelerate training for PLMs, and achieve comparable performance to fine-tuning all parameters, especially at large scales. Therefore, we adopt BitFit~\cite{zaken2021bitfit} implemented by OpenDelta\footnote{\url{https://github.com/thunlp/OpenDelta}} to fine-tune big models. 

\begin{table}
    \centering
    \small
    \begin{tabular}{llll}
    \toprule
    Model & CSJ & CPJ & CiC \\ 
    \midrule
    \BERTbase & Query Entity & Concept & All \\ 
    \Rbase & Query Entity & Concept & Concept \\ 
    \GPTbase& All & All & Concept \\ 
    \GPTNeobase & All & Concept & Concept \\ 
    \BARTbase & Query Entity & Concept & Concept \\ 
    \Tbase & Query Entity & Concept & All \\ 
    \bottomrule
    \end{tabular}
    \caption{The text parts used to calculate scores of prompts in zero-shot probing on the three datasets.
    \textbf{All}: use the negative perplexities of prompts as scores.
    The meanings of the other text parts are shown in \cref{fig:method}.}
    \label{tab:appendix_textspan}
\end{table}

\begin{table*}
    \centering
    \small
    \resizebox{\linewidth}{!}{%
    \begin{tabular}{l}
    \toprule
    Conceptual Similarity Judgment\\
    \midrule
    \textbf{Original Query}: Inter Milan \\
    \textbf{Original Candidates}: Milan, Milan Fashion Week, Pohang Steelers, Series A \\
    \textbf{Original Label}: Pohang Steelers  \\
    
    \textbf{Processed Input}: choose the most similar entity to Inter Milan: (A) Milan, (B) Milan Fashion Week, (C) Pohang Steelers, (D) Series A. \\ 
    \textbf{Processed Label}: C \\ 
    \midrule
    Conceptual Property Judgment \\ 
    \midrule
    \textbf{Original Statement}: Mammals raise their young on milk. \\
    \textbf{Original Label}: True \\ 
    \textbf{Processed Input}: verify: Mammals raise their young on milk. \\
    \textbf{Processed Label}: true \\ 
    \midrule
    Conceptualization in Contexts\\
    \midrule
    \textbf{Original Context}: \textit{Dolly} is running on the grassland. \\
    \textbf{Concept Chain}: Horse --> Mammal --> Animal \\ 
    \textbf{Original Label}: Animal  \\
    \textbf{Processed Input}: select concept: <entity> Dolly </entity> is running on the grassland. Select a contextually related concept for \\ 
    \hspace{16ex} Dolly from (A) Horse, (B) Mammal, (C) Animal. \\
    \textbf{Processed Label}: C \\ 
    \bottomrule
    \end{tabular}
    }
    \caption{The input and output format
    used to linear probe and fine-tune T5 on the three datasets.}
    \label{tab:appendix_t5}
\end{table*}

\begin{table*}
    \centering
    \small
    \begin{tabular}{l|rrcrrcrr}
    \toprule
    & \multicolumn{2}{c}{CSJ}  &
    \phantom{ab} & 
    \multicolumn{2}{c}{CPJ} & 
    \phantom{ab} & 
    \multicolumn{2}{c}{CiC} \\
    & The Others & T5 & & The Others & T5 & & The Others & T5 \\ 
    \midrule
    Learning Rate & $3 \times 10^{-5}$ & $5 \times 10^{-5}$ & & $3 \times 10^{-5}$ & $5 \times 10^{-5}$ & & $3 \times 10^{-5}$ & $5 \times 10^{-5}$ \\
    Weight Decay & $1 \times 10^{-5}$ & $1 \times 10^{-5}$ & & $1 \times 10^{-5}$ & $1 \times 10^{-5}$ & & $1 \times 10^{-5}$ & $1 \times 10^{-5}$ \\
    Batch Size & $4$ & $16$ & & $64$ & $32$ & & $4$ & $16$ \\
    Warmup Rate & $0.1$ & $0.1$ & & $0.1$ & $0.1$ & & $0.1$ & $0.1$ \\
    \bottomrule
    \end{tabular}
    \caption{Hyperparameters used to fine-tune PLMs on \ourdataset.}
    \label{tab:appendix_hyper}
\end{table*}

\begin{table*}
    \centering
    \small
    \begin{tabular}{l|rrr|rrr|rr}
    \toprule
    & \multicolumn{3}{c|}{CSJ} & 
    \multicolumn{3}{c|}{CPJ} &
    \multicolumn{2}{c}{CiC} \\ 
    & Query Entity & Candidate Entity & All & Concept & Answer & All & Concept & All \\
    \midrule
\BERTsmall&$15.0$&$6.5$&$8.1$&$50.7$&$48.5$&$51.5$&$31.9$&$35.1$\\
\BERTmedium&$16.8$&$7.2$&$10.0$&$49.3$&$46.7$&$49.2$&$29.6$&$33.3$\\
\BERTbase&$20.3$&$7.5$&$11.3$&$49.4$&$47.2$&$49.2$&$32.6$&$37.6$\\
\BERTlarge&$22.3$&$8.2$&$13.4$&$50.5$&$47.6$&$50.4$&$31.1$&$36.9$\\
\Rbase&$15.5$&$5.1$&$10.0$&$49.2$&$46.7$&$47.6$&$31.4$&$25.5$\\
\GPTbase&$2.9$&$6.6$&$7.9$&$49.4$&$48.4$&$51.5$&$32.3$&$31.1$\\
\GPTmedium&$3.7$&$8.6$&$10.5$&$52.0$&$47.2$&$47.2$&$30.3$&$32.0$\\
\GPTlarge&$4.6$&$9.0$&$11.3$&$51.8$&$47.3$&$47.2$&$34.3$&$33.8$\\
\GPTxl&$3.9$&$9.6$&$11.7$&$50.7$&$47.2$&$47.1$&$35.3$&$37.0$\\
\GPTNeobase&$2.6$&$6.6$&$7.9$&$52.2$&$47.2$&$47.6$&$32.6$&$28.8$\\
\BARTbase&$14.4$&$5.0$&$7.1$&$48.7$&$48.4$&$48.0$&$33.6$&$27.4$\\
\Tsmall&$11.6$&$5.4$&$6.5$&$52.5$&$47.6$&$53.2$&$34.9$&$40.1$\\
\Tbase&$15.2$&$7.2$&$10.3$&$55.9$&$47.2$&$49.5$&$39.1$&$42.3$\\
\Tlarge&$20.9$&$7.8$&$14.0$&$52.4$&$47.2$&$49.8$&$40.5$&$42.6$\\
\Txl&$19.2$&$7.9$&$14.1$&$49.4$&$47.7$&$49.4$&$38.6$&$47.0$\\
\Txxl&$24.8$&$7.8$&$14.5$&$46.7$&$46.7$&$49.9$&$37.2$&$41.3$\\

    \bottomrule
    \end{tabular}
    \caption{Overall zero-shot probing accuracies (\%) of using different text parts to score prompts on \ourdataset.}
    \label{tab:appendix_probing}
\end{table*}

\begin{table*}
    \centering
    \small
    \resizebox{\linewidth}{!}{%
    \begin{tabular}{l|rrrrr|rrrrr}
    \toprule
    \multirow{2}{*}{Model} & \multicolumn{5}{c|}{Linear Probing} & \multicolumn{5}{c}{Fine-tuning} \\ 
    & Seed=42 & Seed=43 & Seed=44 & Mean & Std & Seed=42 & Seed=43 & Seed=44 & Mean & Std  \\ 
    \midrule
    \multicolumn{11}{c}{Conceptual Similarity Judgment} \\
    \midrule
\BERTsmall&$9.1$&$8.2$&$8.9$&$8.7$&$0.37$&$17.6$&$17.1$&$19.2$&$18.0$&$0.91$\\
\BERTmedium&$13.1$&$12.3$&$13.1$&$12.8$&$0.35$&$20.3$&$21.1$&$21.6$&$21.0$&$0.57$\\
\BERTbase&$16.3$&$16.3$&$15.8$&$16.1$&$0.21$&$28.5$&$26.6$&$26.9$&$27.3$&$0.86$\\
\BERTlarge&$16.5$&$16.9$&$17.3$&$16.9$&$0.31$&$28.7$&$30.2$&$29.5$&$29.5$&$0.61$\\
\Rbase&$11.8$&$12.0$&$12.3$&$12.0$&$0.21$&$22.8$&$21.6$&$22.4$&$22.3$&$0.51$\\
\GPTbase&$4.6$&$4.1$&$4.1$&$4.3$&$0.24$&$19.7$&$20.1$&$20.3$&$20.1$&$0.23$\\
\GPTmedium&$5.3$&$5.2$&$5.2$&$5.2$&$0.02$&$24.9$&$22.2$&$23.0$&$23.4$&$1.15$\\
\GPTlarge&$4.0$&$6.8$&$5.6$&$5.5$&$1.13$&$22.2$&$24.0$&$23.4$&$23.2$&$0.77$\\
\GPTxl&$7.8$&$15.0$&$10.1$&$11.0$&$3.00$&$25.9$&$24.2$&$25.7$&$25.3$&$0.75$\\
\GPTNeobase&$11.1$&$10.7$&$11.2$&$11.0$&$0.20$&$18.8$&$18.4$&$17.8$&$18.3$&$0.42$\\
\BARTbase&$8.5$&$8.3$&$8.4$&$8.4$&$0.10$&$20.4$&$21.0$&$21.7$&$21.0$&$0.50$\\
\Tsmall&$4.8$&$4.8$&$4.7$&$4.8$&$0.05$&$10.1$&$17.6$&$6.9$&$11.5$&$4.48$\\
\Tbase&$5.2$&$4.8$&$4.7$&$4.9$&$0.21$&$27.4$&$27.5$&$28.7$&$27.9$&$0.60$\\
\Tlarge&$4.7$&$4.9$&$4.8$&$4.8$&$0.09$&$31.0$&$33.4$&$32.5$&$32.3$&$1.01$\\
\Txl&$5.0$&$4.9$&$5.2$&$5.0$&$0.11$&$41.0$&$40.6$&$42.0$&$41.2$&$0.61$\\
\Txxl&$4.7$&$4.7$&$4.7$&$4.7$&$0.01$&$43.7$&$43.6$&$43.8$&$43.7$&$0.08$\\

    \midrule
    \multicolumn{11}{c}{Conceptual Property Judgment} \\
    \midrule
\BERTsmall&$57.8$&$58.8$&$57.8$&$58.1$&$0.47$&$66.3$&$66.5$&$67.2$&$66.7$&$0.39$\\
\BERTmedium&$58.2$&$59.6$&$58.5$&$58.8$&$0.59$&$66.7$&$67.5$&$67.3$&$67.2$&$0.35$\\
\BERTbase&$61.2$&$61.9$&$61.5$&$61.6$&$0.28$&$66.8$&$68.3$&$69.2$&$68.1$&$0.98$\\
\BERTlarge&$61.6$&$61.7$&$59.0$&$60.8$&$1.26$&$67.8$&$69.6$&$71.2$&$69.5$&$1.41$\\
\Rbase&$61.7$&$62.0$&$61.9$&$61.9$&$0.13$&$71.4$&$72.7$&$71.8$&$72.0$&$0.54$\\
\GPTbase&$65.2$&$63.3$&$66.0$&$64.8$&$1.14$&$71.3$&$69.5$&$70.5$&$70.4$&$0.72$\\
\GPTmedium&$67.0$&$67.4$&$67.4$&$67.3$&$0.17$&$73.0$&$68.6$&$72.9$&$71.5$&$2.07$\\
\GPTlarge&$66.2$&$67.8$&$66.8$&$66.9$&$0.62$&$74.5$&$72.7$&$73.4$&$73.5$&$0.74$\\
\GPTxl&$67.8$&$68.1$&$68.6$&$68.2$&$0.36$&$74.5$&$75.1$&$74.7$&$74.8$&$0.22$\\
\GPTNeobase&$61.9$&$62.4$&$62.1$&$62.2$&$0.21$&$68.9$&$68.4$&$67.4$&$68.2$&$0.62$\\
\BARTbase&$58.8$&$58.2$&$58.7$&$58.5$&$0.27$&$68.5$&$69.2$&$67.1$&$68.2$&$0.86$\\
\Tsmall&$67.7$&$67.2$&$65.0$&$66.6$&$1.18$&$71.3$&$72.2$&$72.1$&$71.9$&$0.40$\\
\Tbase&$67.3$&$66.8$&$66.8$&$66.9$&$0.25$&$72.6$&$72.1$&$72.8$&$72.5$&$0.28$\\
\Tlarge&$68.9$&$69.7$&$69.3$&$69.3$&$0.33$&$72.5$&$73.4$&$75.2$&$73.7$&$1.10$\\
\Txl&$69.2$&$69.7$&$69.5$&$69.5$&$0.22$&$76.6$&$76.6$&$76.2$&$76.4$&$0.19$\\
\Txxl&$67.3$&$66.5$&$66.0$&$66.6$&$0.53$&$78.2$&$78.3$&$79.2$&$78.6$&$0.46$\\

    \midrule
    \multicolumn{11}{c}{Conceptualization in Contexts} \\
    \midrule
\BERTsmall&$32.4$&$32.7$&$33.3$&$32.8$&$0.38$&$44.6$&$47.0$&$48.4$&$46.6$&$1.55$\\
\BERTmedium&$31.6$&$31.2$&$31.1$&$31.3$&$0.22$&$49.4$&$49.1$&$49.8$&$49.4$&$0.31$\\
\BERTbase&$33.6$&$34.5$&$35.0$&$34.3$&$0.59$&$49.3$&$48.9$&$50.3$&$49.5$&$0.60$\\
\BERTlarge&$35.4$&$38.9$&$35.3$&$36.6$&$1.67$&$50.7$&$53.0$&$51.6$&$51.8$&$0.92$\\
\Rbase&$27.3$&$32.0$&$30.7$&$30.0$&$1.98$&$51.3$&$52.6$&$53.8$&$52.6$&$1.02$\\
\GPTbase&$31.7$&$36.7$&$35.1$&$34.5$&$2.08$&$54.0$&$54.2$&$54.3$&$54.2$&$0.12$\\
\GPTmedium&$29.3$&$25.6$&$29.1$&$28.0$&$1.69$&$54.6$&$54.5$&$54.9$&$54.7$&$0.14$\\
\GPTlarge&$32.8$&$28.8$&$33.7$&$31.8$&$2.16$&$53.4$&$52.7$&$53.6$&$53.3$&$0.36$\\
\GPTxl&$27.7$&$32.2$&$29.9$&$29.9$&$1.83$&$52.6$&$54.4$&$54.4$&$53.8$&$0.88$\\
\GPTNeobase&$38.9$&$38.9$&$40.9$&$39.6$&$0.93$&$47.6$&$47.0$&$47.5$&$47.4$&$0.25$\\
\BARTbase&$44.1$&$42.1$&$44.9$&$43.7$&$1.19$&$50.8$&$49.7$&$53.5$&$51.3$&$1.56$\\
\Tsmall&$25.7$&$26.1$&$24.9$&$25.6$&$0.53$&$43.5$&$44.4$&$45.0$&$44.3$&$0.64$\\
\Tbase&$25.5$&$23.9$&$24.7$&$24.7$&$0.66$&$53.2$&$53.3$&$52.9$&$53.2$&$0.18$\\
\Tlarge&$24.3$&$24.3$&$25.3$&$24.6$&$0.49$&$52.4$&$56.9$&$57.2$&$55.5$&$2.21$\\
\Txl&$26.7$&$27.5$&$26.8$&$27.0$&$0.35$&$59.2$&$57.5$&$55.9$&$57.5$&$1.35$\\
\Txxl&$25.1$&$26.6$&$26.4$&$26.0$&$0.66$&$56.7$&$58.7$&$56.5$&$57.3$&$0.97$\\

    \bottomrule
    \end{tabular}
    }
    \caption{Overall linear probing and fine-tuning accuracies (\%) of all PLMs on COPEN. We run experiments $3$ times using three seeds: $42$, $43$, $44$. Mean: mean accuracy of the three trials; Std: standard deviation.}
    \label{tab:appendix_overallft}
\end{table*}

\section{More Discussions on Experimental Results}
\label{sec:appendix_discuss}
In the section, we discuss some detailed and interesting observations. 
\paragraph{Comparison of Pre-training Method}
In Figure~\ref{tab:main_experiment}, we can observe that: (1) For PLMs using the same architecture, T5 generally outperforms BART, and BERT generally outperforms RoBERTa. The differences may come from the different pre-training corpora. (2) Autoregressive LMs (GPT-2, GPT-Neo) perform worse on CSJ, which is consistent with the observations on factual knowledge probing~\citep{p-tuning}. As we are the first to study conceptual knowledge in PLMs, we focus on the general question ``to what extent do current PLMs understand conceptual knowledge?'' and provide more general conclusions in the paper. We leave the detailed and in-depth analysis of a specific PLM, e.g., layer-wise analysis~\citep{dalvi2021discovering}, in future works.
\paragraph{Comparison of Probing Method}
Intuitively, zero-shot probing reflects the \textit{lower bound} of PLMs' knowledge~\citep{DBLP:journals/tacl/JiangXAN20}, while linear probing learns a task-specific linear classifier and performs better than zero-shot probing, and fine-tuning reflects the \textit{upper bound} of PLMs' knowledge. However, as shown in  Figure~\ref{tab:main_experiment}, linear probing sometimes underperforms zero-shot probing, especially in CSJ and chain-level CPJ. The reason may be that the concepts used for training and testing are disjoint, and linear probing involves trainable parameters, which may learn spurious or shallow correlations on training sets and hence struggles on generalization. Meanwhile, fine-tuning still performs poorly, which demonstrates that existing PLMs systematically lack conceptual knowledge.
\paragraph{Comparison of Instance-Level and Chain-Level CPJ}
For chain-level, BERT performs the best, but for instance-level performs worse than T5. The reason may be that BERT better understands concept transitivity (i.e., making more consistent predictions) but stores fewer conceptual properties overall. A thorough and comprehensive analysis is needed on this phenomenon and we leave it in future works.

\section{Additional Experimental Results}
\label{sec:appendix_exp}
\cref{tab:appendix_probing} shows overall zero-shot probing results on \ourdataset.
The experimental results of linear probing and fine-tuning are obtained at $3$ random trials using seeds $42$, $43$, $44$. \cref{tab:appendix_overallft} shows overall linear probing and fine-tuning results on \ourdataset. And we provide additional results for the analytical experiments: analysis of \textit{conceptual hallucination} on the CPJ dataset (\cref{sec:appendix_hallucination}), error analysis on the CiC dataset (\cref{sec:appendix_erroroncic}),
and analysis on avoiding dataset artifacts (\cref{sec:analyze_artifact}).

\begin{table}
    \centering
    \small
    \begin{tabular}{lrr}
    \toprule
    Model & Disambiguation & Wrong Level \\ 
    \midrule
    \BERTbase & $29.0\%$ & $71.0\%$\\
    \Rbase &$12.8\%$ &$87.2\%$ \\
    \GPTbase & $12.5\%$ & $87.5\%$ \\
    \GPTNeobase &$11.9\%$ & $88.1\%$\\
    \BARTbase & $11.5\%$& $88.5\%$\\
    \Tbase & $32.0\%$ & $68.0\%$\\ 
    \bottomrule
    \end{tabular}
    \caption{The proportion of different error types of  zero-shot probing results on the CiC dataset. We only consider the entities with more than one concept chain.}
    \label{tab:task3_erroranalysis}
\end{table}

\subsection{\textit{Conceptual Hallucination} on CPJ}
\label{sec:appendix_hallucination}
\begin{figure*}
    \centering
    \includegraphics[width=\linewidth]{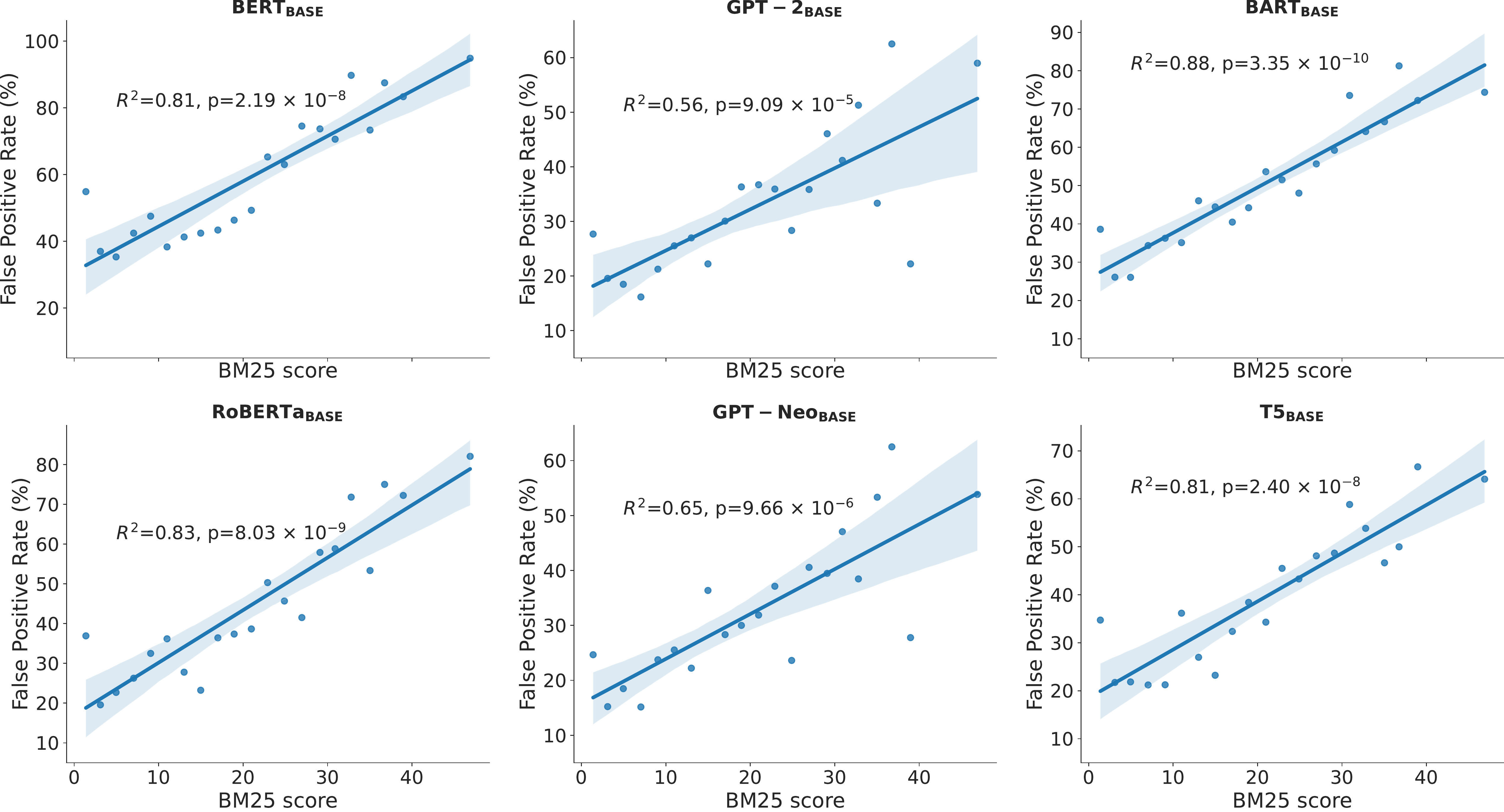}
    \caption{The false positive rate of various PLMs' fine-tuning results on negative instances of the CPJ dataset with different BM25 scores.}
    \label{fig:bm25}
\end{figure*}

Figure~\ref{fig:bm25} shows the false negative rates on subsets with different BM25 scores for various PLMs. We can observe that
the false positive rates, which indicates conceptual hallucination, have strong positive correlations to the BM25 scores, which indicates word co-occurrence. 

\subsection{Error Analysis on CiC}
\label{sec:appendix_erroroncic}
Table~\ref{tab:task3_erroranalysis} shows the proportions of different error types. We can observe that in most wrong predictions, PLMs select concepts of wrong levels. It indicates that PLMs lack a comprehensive understanding of concept hierarchy and fail to
conceptualize entities according to contexts.

\subsection{Analysis on Avoiding Dataset Artifacts}
\label{sec:analyze_artifact}

Dataset artifacts leak shallow information and cause the PLMs to learn spurious correlations rather than exhibit inner knowledge. When construct \ourdataset, we avoid two kinds of artifacts:

\paragraph{Lexical Overlap}
means that the query and the answer have word overlap, which may enable PLMs to make correct predictions using spurious correlations without the correct knowledge. For example, in CSJ, if the query entity is \texttt{Stanford University} and the answer entity is \texttt{University of California}; in CiC, if the context is \textit{She graduated from \texttt{Stanford University}} and the answer concept is \texttt{University}; they have lexical overlap. 

We conduct experiments on the data with lexical overlap. As shown in \cref{tab:trivial}, on the data with lexical overlap, PLMs perform much better. But this should be interpreted as they learn shallow clues leaked by artifacts since they cannot achieve similar performance on data without lexical overlap. Hence, we filter out all instances with lexical overlap in \ourdataset to avoid this kind of artifact. 

\begin{table}[t!]
    \centering
    \small
    \resizebox{\columnwidth}{!}{%
    \begin{tabular}{l|rr|rr}
    \toprule
    \multirow{2}{*}{Model} &
    \multicolumn{2}{c|}{CSJ} &
    \multicolumn{2}{c}{CiC} \\
    & w/ LO & w/o LO & w/ LO & w/o LO \\
    \midrule
    \BERTbase&$68.9$&$20.3$&$52.5$&$37.6$\\
    \Rbase&$62.2$&$15.5$&$48.5$&$31.4$\\
    \GPTbase&$34.2$&$7.9$&$43.8$&$32.3$\\
    \GPTNeobase&$34.0$&$7.9$&$52.4$&$32.6$\\
    \BARTbase&$75.9$&$14.4$&$53.2$&$33.6$\\
    \Tbase&$69.2$&$15.2$&$62.7$&$42.3$\\
    \bottomrule
    \end{tabular}
    }
    \caption{Zero-shot probing accuracies (\%) of PLMs on 
    data with lexical overlap (w/ LO) and without lexical overlap (w/o LO). We collect $688$ and $1,200$ instances with lexical overlap for CSJ and CiC, respectively.}
    \label{tab:trivial}
\end{table}

\paragraph{Concept Overlap} is that the same concepts show up in both training and test datasets, which may leak conceptual knowledge, i.e., the PLMs may learn some knowledge from training data. In \ourdataset, as mentioned in \cref{sec:concept_taxonomy}, we split different top-level concepts and their subconcepts into different sub-datasets, so as to avoid concept overlap. To empirically show the influence of concept overlap, we randomly re-split the datasets into same-size training, development, and test sets and see the fine-tuning performance on the new split.

\begin{figure}[t!]
    \centering
    \includegraphics[width=0.80\linewidth]{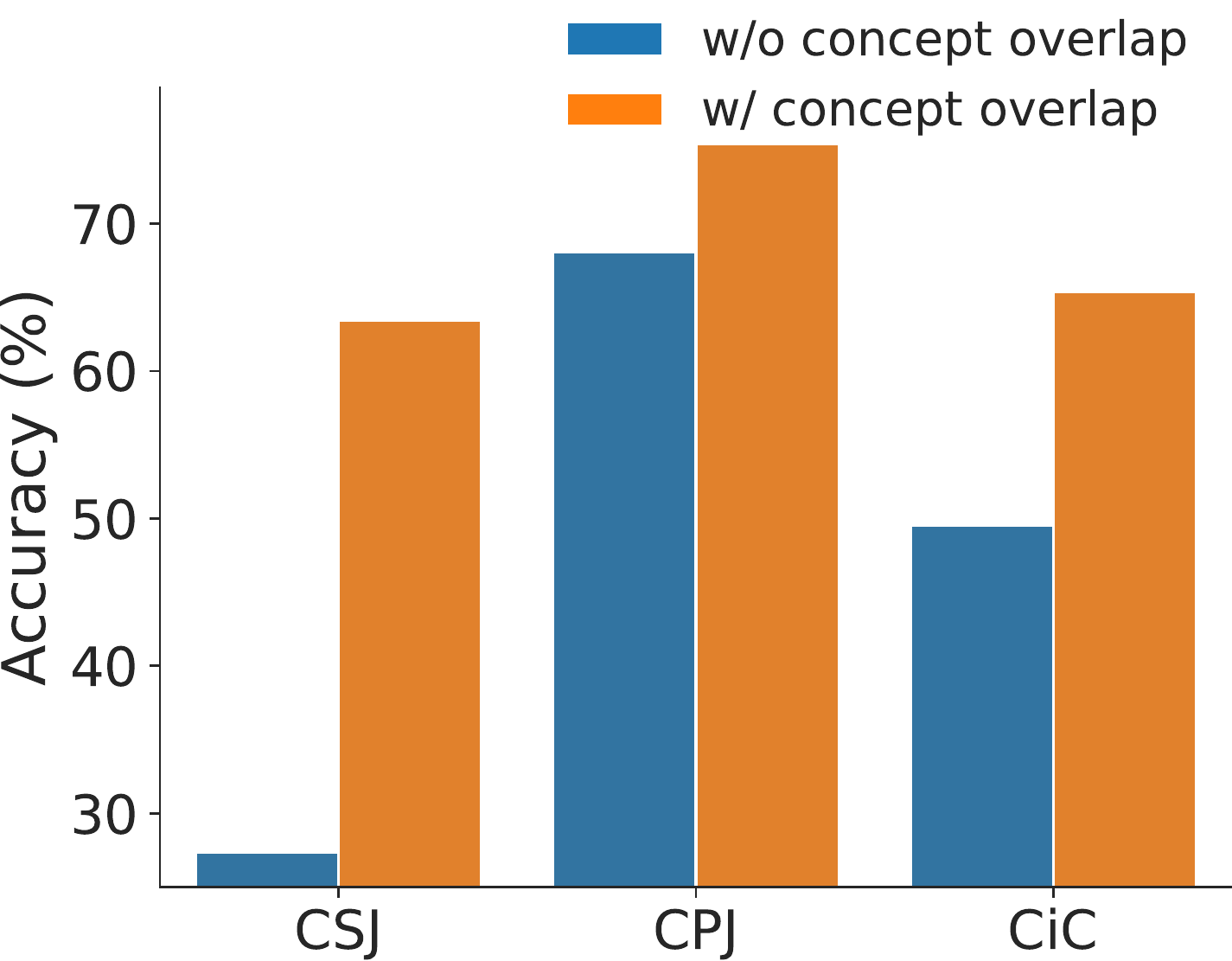}
    \caption{Fine-tuning accuracies of \BERTbase on data with and without concept overlap. }
    \label{fig:concept_overlap}
\end{figure}

The results of fine-tuning BERT are shown in \cref{fig:concept_overlap}, and the results of fine-tuning and linear probing for all PLMs are shown in Table~\ref{tab:appendix_conceptoverlap}. Fine-tuning on datasets with concept overlap achieves much higher accuracies, especially on CSJ. It indicates that if we do not avoid concept overlap, PLMs can easily learn conceptual knowledge from training data and lead to false optimistic conclusions. 

\begin{table*}
    \centering
    \small
    \begin{tabular}{l|rr|rr|rr}
    \toprule
    \multirow{2}{*}{Model}  & \multicolumn{2}{c|}{CSJ} & \multicolumn{2}{c|}{CPJ} & \multicolumn{2}{c}{CiC} \\ 
    & w/ CO & w/o CO &  w/ CO & w/o CO &  w/ CO & w/o CO \\ 
    \midrule
\multicolumn{7}{c}{Linear Probing} \\
\midrule
\BERTbase&$20.0$&$16.1$&$64.1$&$61.6$&$46.5$&$34.3$\\
\Rbase&$12.3$&$12.0$&$65.9$&$61.9$&$45.4$&$30.0$\\
\GPTbase&$5.2$&$4.3$&$67.2$&$64.8$&$39.0$&$34.5$\\
\GPTNeobase&$15.4$&$11.0$&$64.6$&$62.2$&$58.3$&$39.6$\\
\BARTbase&$9.4$&$8.4$&$62.6$&$58.5$&$50.2$&$43.7$\\
\Tbase&$4.7$&$4.9$&$68.8$&$66.9$&$33.9$&$24.7$\\
    \midrule
    \multicolumn{7}{c}{Fine-tuning} \\ 
    \midrule
\BERTbase&$63.4$&$27.3$&$75.4$&$68.1$&$65.4$&$49.5$\\
\Rbase&$61.0$&$22.3$&$77.0$&$72.0$&$66.6$&$52.6$\\
\GPTbase&$49.9$&$20.1$&$72.7$&$70.4$&$65.4$&$54.2$\\
\GPTNeobase&$44.3$&$18.3$&$71.2$&$68.2$&$62.5$&$47.4$\\
\BARTbase&$54.7$&$21.0$&$73.1$&$68.2$&$67.4$&$51.3$\\
\Tbase&$50.6$&$27.9$&$77.6$&$72.5$&$67.6$&$53.2$\\
\bottomrule
    \end{tabular}
    \caption{Accuracies (\%) of linear probing and fine-tuning on data with concept overlap (w/ CO) and without concept overlap (w/o CO).}
    \label{tab:appendix_conceptoverlap}
\end{table*}

\begin{table*}
    \centering
    \small
    \begin{tabular}{lcl}
    \toprule
    & \#Concepts & Top-Level Concepts \\ 
    \midrule
    Training\& & \multirow{2}{*}{$248$} & Organisation, Name, Award, MeanOfTransportation, Colour, Language, Person, \\
    Development &  & Holiday, Work, Currency, EthnicGroup \\ 
    \midrule
    \multirow{2}{*}{Testing}& \multirow{2}{*}{198} & AnatomicalStructure, Species, Food, Event, TimePeriod, ChemicalSubstance,  \\
    & & 
    Place, Device, Disease, Activity, Biomolecule, SportsSeason\\
    \bottomrule
    \end{tabular}
    \caption{The top-level concepts and the number of concepts used for training, development, and testing. }
    \label{tab:appendix_concepts}
\end{table*}

\section{\ourdataset}
\label{sec:appendix_copen}
We provide a detailed introduction to \ourdataset.
\subsection{\ourdataset Taxonomy}
\label{sec:taxonomy_appendix}

\begin{figure*}
\pagebreak
    \centering
    \includegraphics[width=0.57\linewidth]{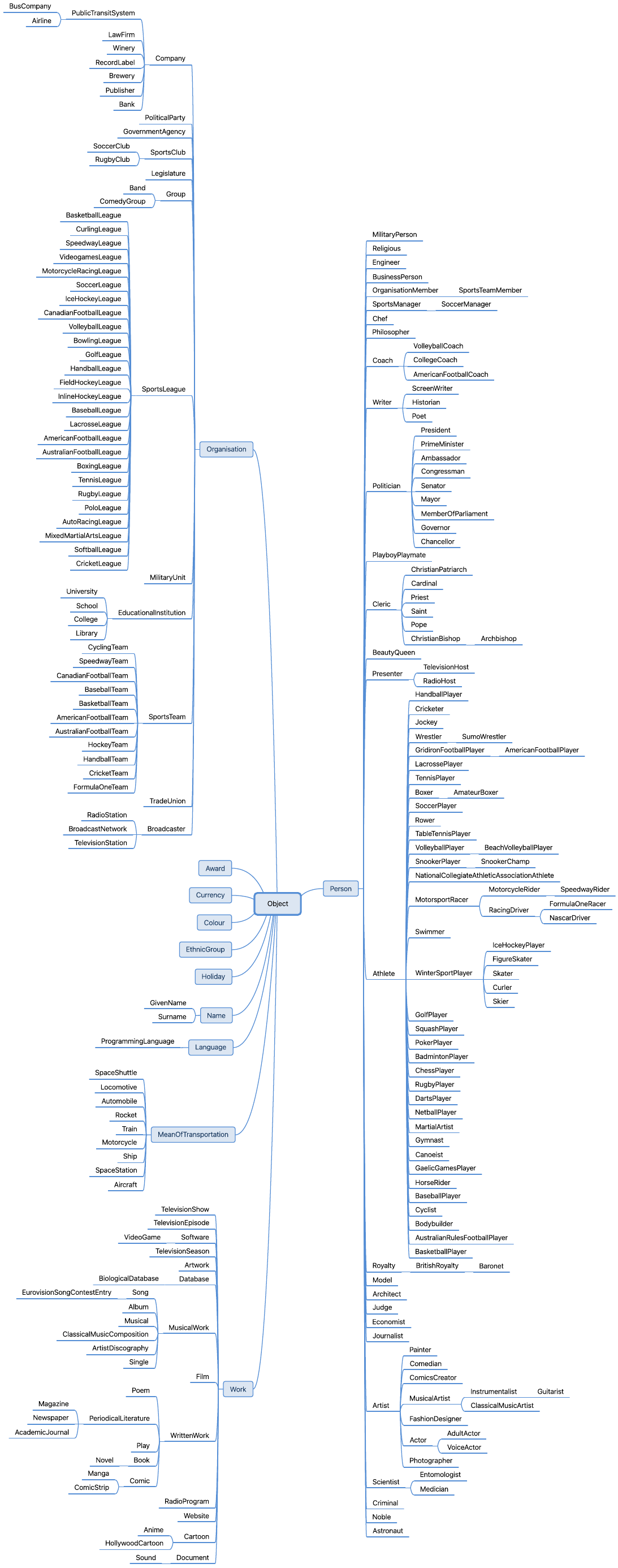}
    \caption{Concept taxonomy for training and development datasets. \texttt{Object} is a virtual concept without annotated instances.}
    \label{fig:appendix_traintaxo}
\end{figure*}

\begin{figure*}
\pagebreak
    \centering
    \includegraphics[width=1.0\linewidth]{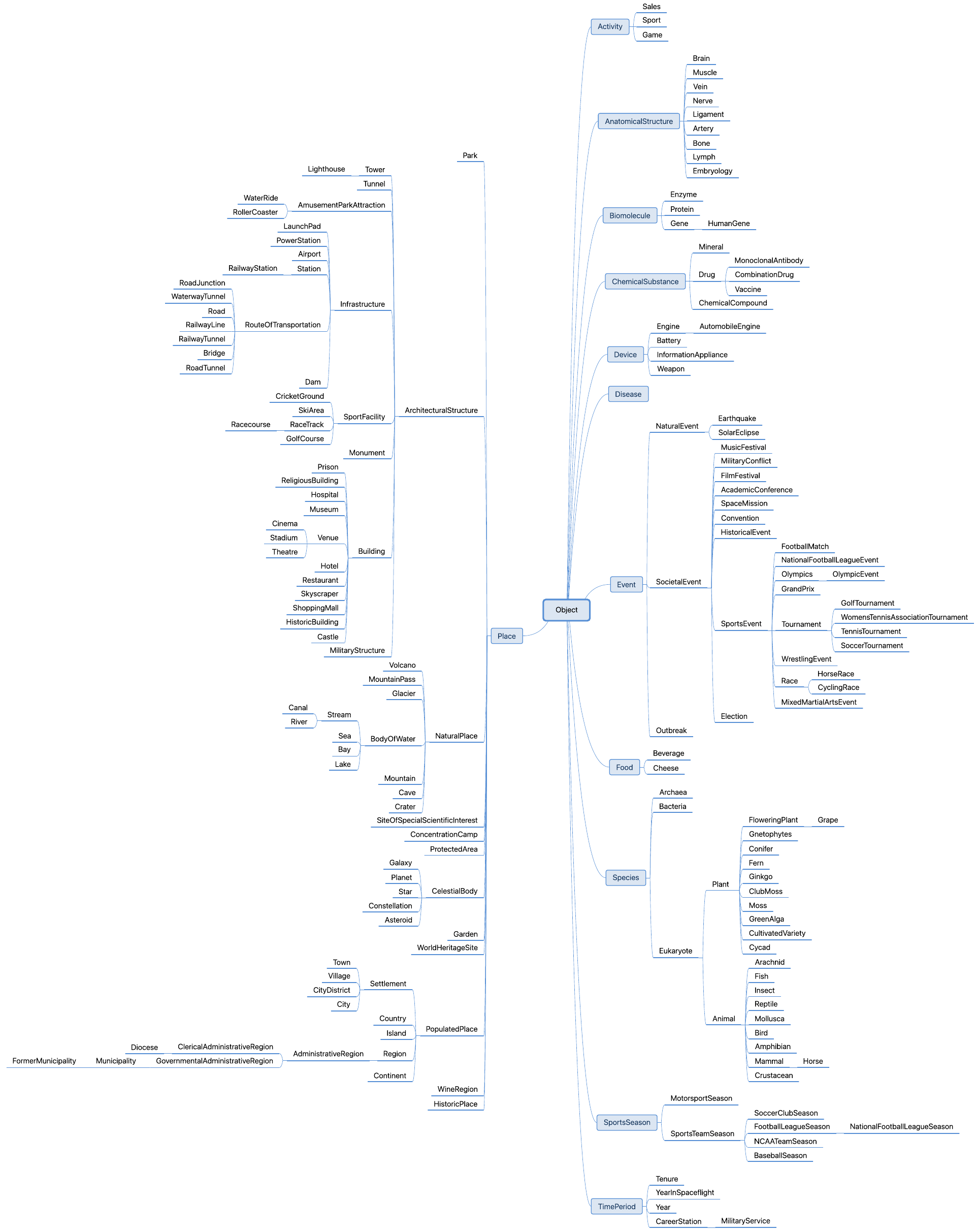}
    \caption{Concept taxonomy for testing datasets. \texttt{Object} is a virtual concept without annotated instances.}
    \label{fig:appendix_testtaxo}
\end{figure*}

\paragraph{Disjoint Concepts}
We divide all the concepts into two disjoint sets:
one set containing $11$ top-level concepts together with all their sub-concepts for 
constructing training and development datasets, and 
the other set containing the other concepts for testing datasets. As shown in Table~\ref{tab:appendix_concepts}, there are $248$ concepts including $11$ top-level concepts for training and development datasets and $198$ concepts including $12$ top-level concepts for testing. 
\paragraph{Concept Hierarchy}
We present the concepts for training and development datasets in \cref{fig:appendix_traintaxo} and the concepts for testing datasets in \cref{fig:appendix_testtaxo}. \texttt{Object} is a virtual concept for visualization and is not included in the overall $446$ concepts.

\subsection{Concept Similarity Judgment}
\label{sec:appendix_CSJ}

\paragraph{Human Performance}

We sample $1,000$ instances from the testing dataset and 
invite annotators with no linguistic background
to perform the CSJ task.
All the annotators are trained with a few instances before the evaluation.

\paragraph{Co-occurrence-based Filtering}
We filter out instances of which query entities and answer entities have a high association, which are estimated by cosine similarity of their Glove word embeddings. Specifically, for a query entity, we sample $5$ answer entities and select the entity with the lowest association with the query entity as the answer entity. Then we choose distractor entities iteratively following the rules: (1) Sample a distractor entity, if the entity has a higher association with the query entity than the answer entity, then select the distractor entity as a candidate entity. (2) If not, select the distractor entity as a candidate entity with a $20\%$ probability, otherwise start the next iteration until the number of distractor entities reaches $20$.

\subsection{Conceptual Property Judgment}
\label{sec:appendix_CPJ}
\paragraph{Human Annotation}
We invite annotators with no linguistics background to check
whether the instances are correctly labeled, grammatically correct, and describing concept properties. All annotators are well-trained and required to pass a qualification before the annotation. 
The instances originally labeled as false are annotated $4$ times, and the other instances are annotated once. 
During the annotation, an author of the
paper and another experienced annotator separately sample 10\% of the instances to check the quality of annotation.

The acceptance criterion of the annotation is that the percentage of obvious annotation errors in the sampled instances (e.g., label the statement \textit{The sun has two eyes} as true) does not exceed 3\%, and the inter-annotator agreement rates exceed 85\% for the instances annotated $4$ times. Major voted results of the instances annotated $4$ times together with 
the instances annotated once constitute
the CPJ dataset.

\paragraph{Human Performance}
We use the $2{,}159$ instances that are annotated $4$ times in the testing dataset to evaluate human performance. We conduct a $4$-round evaluation: take the major voted results of $3$ annotators as labels and the other one as human predictions to calculate the accuracy of the round.
The mean accuracy of $4$
rounds is reported as the human accuracy on the CPJ dataset.

\subsection{Conceptualization in Contexts}
\label{sec:appendix_CiC}
\paragraph{Human Annotation}
We invite annotators with no linguistics background to annotate the dataset.
To ensure quality, all annotators are well-trained and required to pass a qualification before the annotation. All instances are annotated four times.
Moreover, during the annotation, an author of the paper and another experienced annotator separately sample 10\% of the examples to check the quality of annotation. 
The acceptance criterion of the annotation is that the percentage of obvious annotation errors (e.g., Select \texttt{Horse} for \texttt{Dolly} according to the context \textit{Dolly is running on the grassland.}) does not exceed 3\%, and the inter-annotator agreement rates exceed 80\%. 
 Major voted results of the $4$ annotated results constitute the final CiC dataset.

\paragraph{Human Performance}
We use all instances in the testing dataset, which are annotated $4$ times, to evaluate human performance. We conduct a $4$-round evaluation: take the major voted results of $3$ annotators as labels and the other one as human predictions to calculate the accuracy of the round. The mean accuracy of $4$ rounds is the human accuracy.

\end{document}